\pdfoutput=1

\documentclass[11pt]{article}

\usepackage[final]{acl}

\usepackage{makecell}
\usepackage{subcaption}
\usepackage{CJKutf8}
\usepackage{times}
\usepackage{latexsym}
\usepackage{epsfig}
\usepackage{graphicx}
\usepackage{xcolor}
\usepackage{amsmath}
\usepackage{amssymb}
\usepackage{algpseudocode}

\usepackage{multirow}
\usepackage{booktabs}
\usepackage{algorithm}
\usepackage{bbm}
\usepackage{amssymb}
\usepackage{colortbl}
\usepackage{pifont}
\definecolor{mypink}{rgb}{.99,.91,.95}
\definecolor{mygreen}{rgb}{.9,.99,.9}
\definecolor{mygray}{gray}{.9}

\usepackage{tcolorbox}
\usepackage{listings}
\usepackage{enumitem}
\setenumerate[1]{itemsep=-1pt,partopsep=0pt,topsep=0pt}
\setitemize[1]{itemsep=-1pt,partopsep=0pt,topsep=0pt}
\setdescription{itemsep=-1pt,partopsep=0pt,topsep=0pt}

\usepackage[T1]{fontenc}

\usepackage[utf8]{inputenc}

\usepackage{microtype}

\usepackage{inconsolata}

\usepackage{graphicx}

\title{Mixture of Decoding: An Attention-Inspired Adaptive Decoding Strategy to Mitigate Hallucinations in Large Vision-Language Models}

\author{
 \textbf{Xinlong Chen\textsuperscript{1,2}}\thanks{Work done during an internship at Kuaishou Technology.},
 \textbf{Yuanxing Zhang\textsuperscript{3}},
 \textbf{Qiang Liu\textsuperscript{1,2}}\thanks{Corresponding author: \href{qiang.liu@nlpr.ia.ac.cn}{qiang.liu@nlpr.ia.ac.cn}},
 \textbf{Junfei Wu\textsuperscript{1,2}},\\
 \textbf{Fuzheng Zhang\textsuperscript{3}},
 \textbf{Tieniu Tan\textsuperscript{1,2,4}}
\\
 \textsuperscript{1}New Laboratory of Pattern Recognition (NLPR),\\
Institute of Automation, Chinese Academy of Sciences (CASIA)\\
 \textsuperscript{2}School of Artificial Intelligence, University of Chinese Academy of Sciences\\
 \textsuperscript{3}Kuaishou Technology
 \textsuperscript{4}Nanjing University
\\
}

\begin{document}
\maketitle
\begin{abstract}
Large Vision-Language Models (LVLMs) have exhibited impressive capabilities across various visual tasks, yet they remain hindered by the persistent challenge of hallucinations. To address this critical issue, we propose Mixture of Decoding (MoD), a novel approach for hallucination mitigation that dynamically adapts decoding strategies by evaluating the correctness of the model's attention on image tokens. Specifically, MoD measures the consistency between outputs generated from the original image tokens and those derived from the model's attended image tokens, to distinguish the correctness aforementioned. If the outputs are consistent, indicating correct attention, MoD employs a complementary strategy to amplify critical information. Conversely, if the outputs are inconsistent, suggesting erroneous attention, MoD utilizes a contrastive strategy to suppress misleading information. Extensive experiments demonstrate that MoD significantly outperforms existing decoding methods across multiple mainstream benchmarks, effectively mitigating hallucinations in LVLMs. The code is available at \url{https://github.com/xlchen0205/MoD}.
\end{abstract}

\section{Introduction}
Recently, Large Language Models (LLMs)~\cite{touvron2023llama, bai2023qwen} have achieved remarkable success, which has spurred significant research interest in extending them into Large Vision-Language Models (LVLMs)~\cite{liu2024llava, bai2023qwenvl}. These LVLMs demonstrate exceptional performance across various visual tasks. However, despite their impressive versatility, LVLMs are confronted with a critical limitation known as ``hallucination''~\cite{zhou2023analyzing, bai2024hallucination, rawte2023survey, huang2023survey}, a phenomenon in which models generate text that appears semantically coherent, but contains content that either contradicts or lacks grounding in the provided visual information. This fundamental flaw significantly compromises the reliability of LVLMs as trustworthy AI assistants in real-world applications~\cite{wang2023chatcad, he2023survey}, particularly in high-stakes domains such as autonomous driving systems~\cite{chen2024automated, tian2024drivevlm}, where inaccurate interpretation of visual data could lead to severe consequences.

A prominent approach to mitigating hallucinations in LVLMs is developing training-free decoding strategies, particularly contrastive decoding. This method works by contrasting the original logits for next-token prediction with those generated under hallucination scenarios, thereby producing more reliable outputs. Specifically, VCD~\cite{leng2024mitigating} mitigates the language prior bias inherent in the LLM backbone by contrasting the original logits with those generated from Gaussian-noised image. Similarly, M3ID~\cite{favero2024multi} reduces language prior bias by contrasting the original logits with those derived from pure text inputs, strengthening the interaction between visual inputs and model outputs. AvisC~\cite{woo2024don} proposes a novel perspective, suggesting that certain image tokens with excessively high attention weights may trigger hallucinations. Thus, it selects high-attention image tokens from specific layers to obtain hallucination logits for contrastive decoding. 

Despite these advancements, existing methods still exhibit notable limitations. VCD and M3ID primarily attribute hallucinations to language prior bias, neglecting the potential influence of visual inputs, such as spurious correlations~\cite{chen2024multi}. Although AvisC considers the model's attention distribution over the input image, it may erroneously weaken the role of high-attention image tokens when the model has already accurately focused on relevant information, leading to unreliable contrastive results. These limitations underscore a critical gap in current methods: the insufficient consideration of both (1) the influence of visual inputs on hallucination and (2) the variability of the attention distribution, which may be either right or wrong, ultimately restricting their performance in complex scenarios.


\begin{figure}[ht]
  \centering
    \includegraphics[width=\columnwidth]{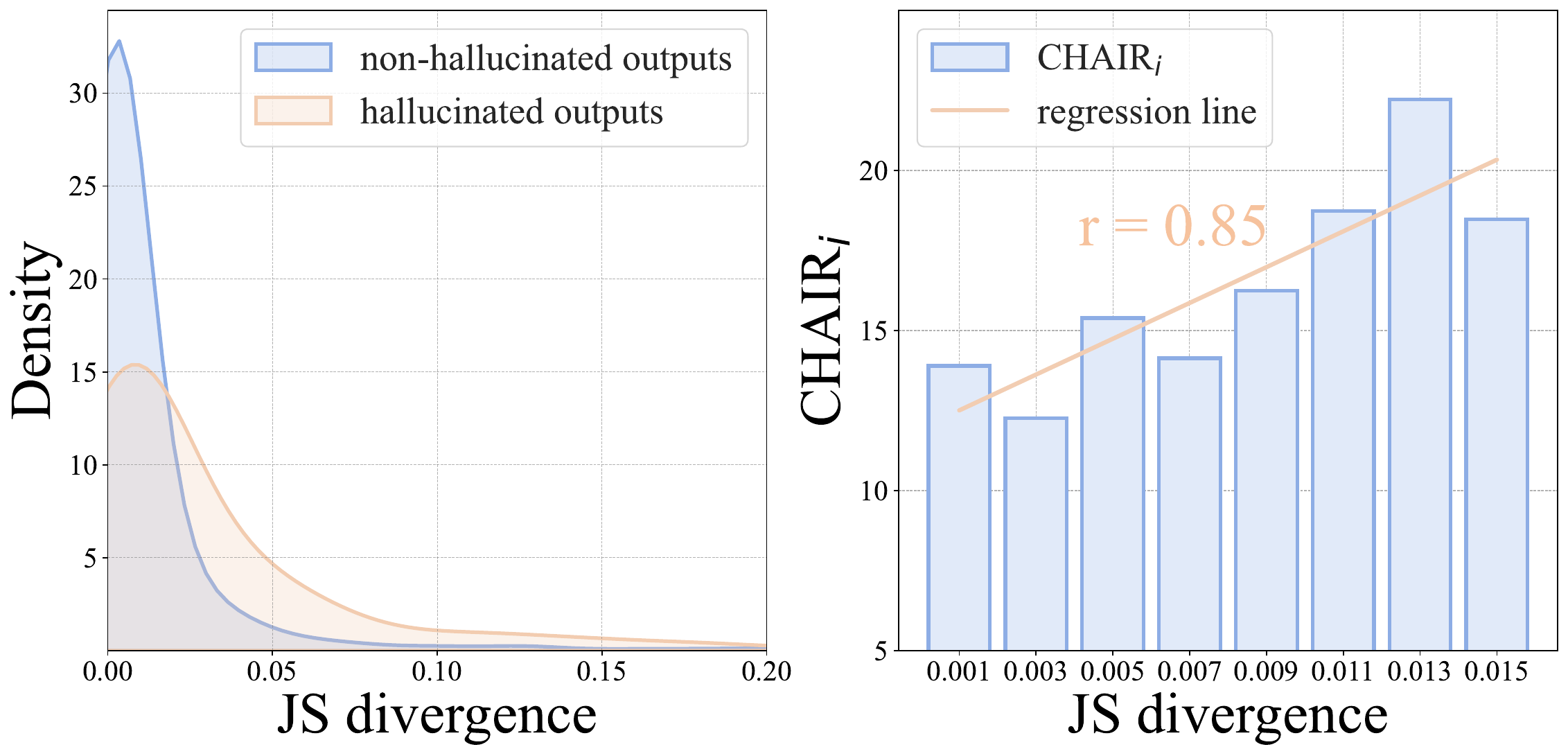}
  \caption{\textbf{The JS divergence excels at distinguishing hallucinated from non-hallucinated outputs}, as evidenced in both discriminative tasks (POPE, left) and generative tasks (CHAIR, right). In the left figure, non-hallucinated outputs are primarily concentrated in regions with lower JS divergence values, while hallucinated outputs exhibit a pronounced long-tail distribution. In the right figure, there is a significant positive correlation between the \(\text{CHAIR}_i\) metric and JS divergence, with a Pearson correlation coefficient of 0.85 (p < 0.01).}
  \label{fig:edxample}
\end{figure}

The key to addressing these limitations lies in assessing the correctness of the model's attention on image tokens and dynamically adjusting the decoding strategy accordingly. If the model's attention is correct, indicating that the attended image tokens are relevant and beneficial, amplifying their information can enhance the output probability of proper tokens; conversely, if the model's attention is incorrect, suggesting that the attended image tokens are irrelevant or misleading, suppressing erroneous information is crucial to counteract hallucinations. Inspired by methods that judge correctness based on consistency~\cite{manakul2023selfcheckgpt, liu2025attention}, we observe that measuring the consistency between the original output logits and the logits derived from the model's attended image tokens using Jensen-Shannon (JS) divergence, can effectively distinguish hallucinated outputs from non-hallucinated ones (as illustrated in Fig.~\ref{fig:edxample}). This approach enables us to determine the correctness of the model's attention on image tokens and adaptively select decoding strategies to effectively mitigate hallucinations.

Building on these insights, we propose \textbf{Mixture of Decoding (MoD)}, a method that adaptively selects decoding strategies by evaluating the consistency between the original output and the output derived from the model's attended image tokens. Specifically, if the outputs are consistent, it indicates that the model's attention on image tokens is correct, and we complement the two to amplify critical information. Conversely, if the outputs are inconsistent, it suggests that the model's attention on image tokens is incorrect, and we contrast the two to suppress misleading information derived from erroneous attention. Extensive experiments on multiple mainstream benchmarks demonstrate that MoD significantly outperforms existing methods in mitigating hallucinations.

The main contributions of this paper are summarized as follows:

\begin{itemize}[leftmargin=*]
    \item We demonstrate that the consistency between the outputs derived from the original image tokens and the model's attended image tokens provides a robust criterion for distinguishing hallucinations.
    \item We propose MoD, a novel method that dynamically selects decoding strategies by evaluating the aforementioned consistency, significantly reducing hallucinations in LVLMs.
    \item Our method achieves state-of-the-art performance across multiple prominent benchmarks, highlighting the efficacy and superiority of MoD.
\end{itemize}

\section{Related Work}
\noindent\textbf{Large vision-language models.}
Inspired by the success of LLMs~\cite{touvron2023llama, bai2023qwen}, recent research has extended their capabilities into the multimodal domain, giving rise to the developments of LVLMs~\cite{liu2024llava, bai2023qwenvl}. By integrating textual instructions with visual inputs, LVLMs are capable of understanding and generating diverse content in a more comprehensive manner. Typically, LVLMs undergo two training stages: feature alignment pre-training and instruction-based fine-tuning. Additionally, recent efforts have attempted to enhance the alignment of model responses with human values through reinforcement learning from human feedback (RLHF)~\cite{sun2023aligning} and preference fine-tuning~\cite{zhou2024aligning}. Despite these achievements, existing LVLMs still struggle with significant hallucination issues, which severely undermine their reliability and stability in practical applications. The MoD approach proposed in this paper offers a promising solution to effectively mitigate hallucinations in LVLMs.

\noindent\textbf{Hallucination mitigation methods in LVLMs.}
Hallucination mitigation methods can be systematically categorized into three major paradigms: pre-inference intervention, in-inference regulation, and post-inference correction. In terms of pre-inference intervention, researchers primarily employ specifically curated anti-hallucination data to perform supervised fine-tuning (SFT) on LVLMs~\cite{jiang2024hallucination, yu2024hallucidoctor}. In-inference regulation strategies involve adjustments to the internal components of LVLMs, including enhancing or suppressing attention heads associated with facts or errors~\cite{li2024inference, yuan2024whispers}, as well as various decoding strategies~\cite{leng2024mitigating, favero2024multi, woo2024don}. Post-inference correction methods mainly detect hallucinations by designing verification questions and retrieve external knowledge for correction~\cite{varshney2023stitch, wu2024logical, yin2024woodpecker, luo2023zero, dhuliawala2023chain}. However, existing methods exhibit significant limitations. Pre-inference intervention approaches rely heavily on large-scale, manually constructed or AI-generated training data, resulting in high costs. In-inference regulation methods, which involve adjusting internal components of the model, require extensive testing on large datasets to identify specific components that need regulation, and may inadvertently impair other capabilities of LVLMs. Post-inference correction methods necessitate multiple inferences to detect hallucinations and depend on external knowledge or tools for correction, significantly increasing inference time. In contrast, contrastive decoding offers a more efficient solution for hallucination mitigation, as it requires no additional training, no large-scale data testing, and no extensive repeated sampling.

\noindent\textbf{Contrastive decoding.}
As a hallucination mitigation method under the in-inference regulation paradigm, contrastive decoding~\cite{li2022contrastive} was initially proposed to leverage the contrast between an expert model and an amateur model, generating outputs superior to those of the expert model alone. Recently, this approach has been extended, with researchers employing various designs to replace the amateur model, intentionally inducing hallucinations and contrasting them with the original outputs to obtain hallucination-free results. For instance, VCD~\cite{leng2024mitigating} amplifies language prior bias by adding Gaussian noise to the input image, while M3ID~\cite{favero2024multi} removes the image input entirely and further amplifies language prior bias using pure text inputs. AvisC~\cite{woo2024don} posits that image patches with excessively high attention weights can trigger hallucinations, while SID~\cite{huo2024self} takes the opposite stance; ICD~\cite{wang2024mitigating} introduces a creative method by designing diverse negative prompts to induce hallucinations; DoLa~\cite{chuang2023dola} suggests that early-layer information, being less mature, can be used to induce hallucinations; ID~\cite{kim2024instructive} employs noisy instructions to induce hallucinations; and IBD~\cite{zhu2024ibd} fine-tunes an image-biased LVLM by amplifying attention to visual tokens and contrasts it with the original LVLM. DeGF~\cite{zhang2025self} proposes a novel approach that first generates an image from text based on the original output, then adaptively selects decoding strategies according to the consistency between the output derived from the generated image and the original output, achieving remarkable hallucination detection performance. Although these methods exhibit certain effectiveness, they often fail to adequately address the impact of visual inputs on hallucinations or overlook the variability of attention distribution, which can be either correct or incorrect. To overcome these limitations, we propose MoD, which adaptively selects decoding strategies based on the consistency between outputs derived from the model's attended image tokens and its original outputs, offering a more promising solution for hallucination mitigation.

\section{Method}

\begin{figure*}[t]
  \includegraphics[width=\linewidth]{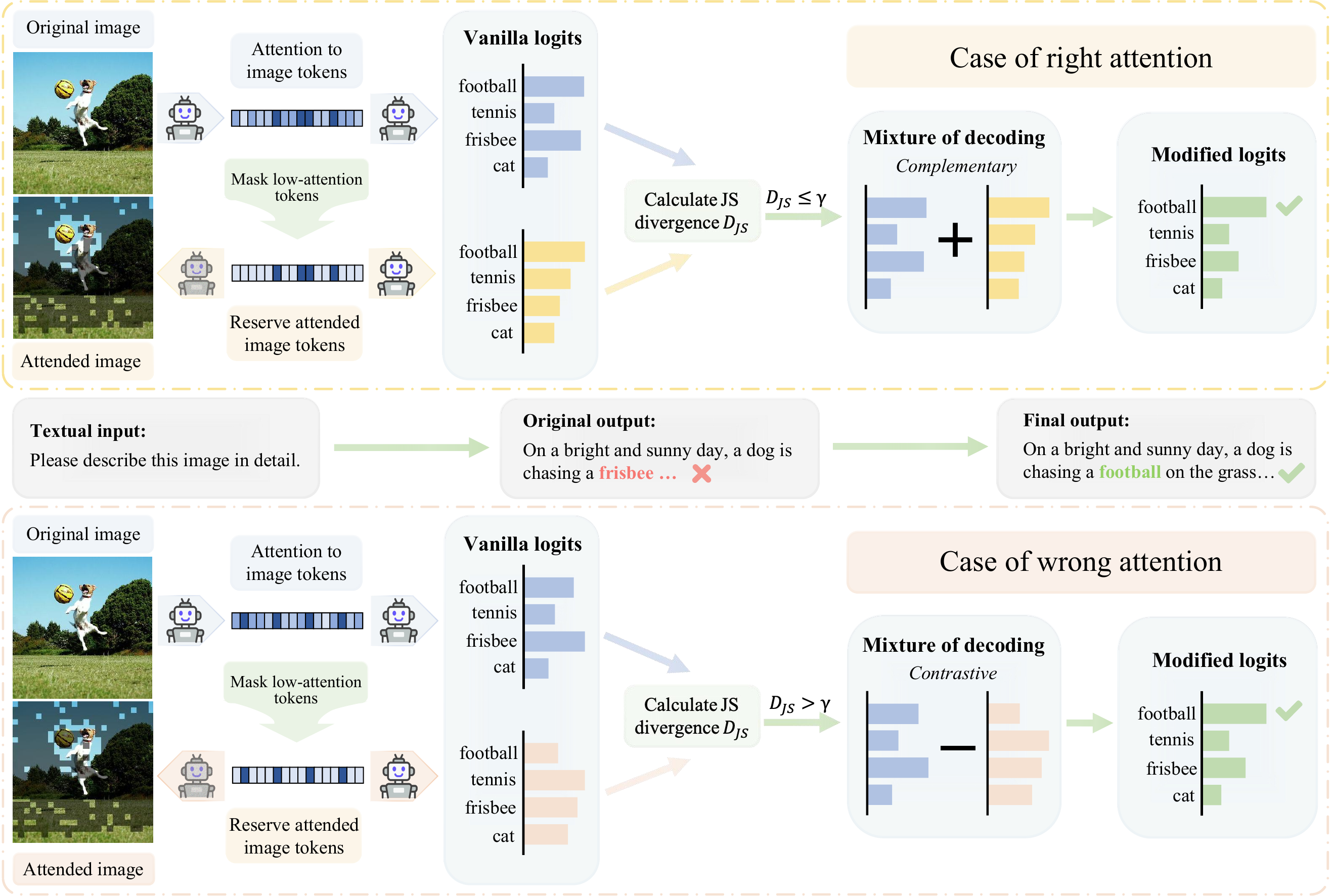}
  \caption {\textbf{Overview of our proposed MoD.} MoD involves three key steps: (1) extracting the model's attended image tokens while masking the others; (2) generating vanilla output logits from both original image tokens and masked image tokens; and (3) computing the JS divergence between the two logit distributions to assess the correctness of the model's attention. Based on this evaluation, MoD adaptively adopts either complementary or contrastive decoding strategies to produce hallucination-free outputs. The upper and lower panels illustrate cases of right and wrong attention, respectively. The corresponding attended image, visualized using LLaVA-1.5, is displayed in the lower-left corner of each panel.}
  \label{fig:overview}
\end{figure*}

\subsection{Preliminary: Attention Formulation}

In the transformer architecture, the attention mechanism plays a pivotal role in the model's decision-making process. Let's begin by defining an embedding sequence \( X \in \mathbb{R}^{N \times d} \), where \( N \) is the total number of tokens, comprising \( N_I \) image tokens and \( N_T \) text tokens such that \( N = N_I + N_T \), and \( d \) represents the dimension of hidden states. This sequence is projected into three distinct matrices: the query \( Q \), the key \( K \), and the value \( V \), through three learnable linear transformations: \( W_Q \), \( W_K \), and \( W_V \). The attention weight matrix \( A \) is then computed as follows:
\begin{equation}
A = \text{softmax}\left(\frac{Q \cdot K^T}{\sqrt{d}}\right).
\end{equation}
The output of the attention mechanism \( O_A \) is derived by \( O_A = A \cdot V \). Let \( A_l \in \mathbb{R}^{B \times H \times N \times N} \) represent the attention weight matrix of layer \( l \) in the LLM backbone consisting of \( L \) layers of an LVLM, where \( B \) and \( H \) denote the batch size and the number of attention heads, respectively.

\subsection{Extraction of Attended Image Tokens}

Due to the auto-regressive nature of LLMs, it can be posited that the final token of the input sequence encapsulates the model's comprehensive understanding of the input query, signifying its readiness to formulate a response. Therefore, we utilize its attention to the image tokens, denoted as \( A^I \in \mathbb{R}^{B \times N_I} \), to identify the specific image tokens that the model has attended to. Specifically, we compute the average attention weight across all layers and attention heads to integrate the model's interpretation of the input across various levels and dimensions of abstraction, formalized as:
\begin{equation}
    A^I = \frac{1}{L \cdot H} \sum_{l=1}^{L} \sum_{h=1}^{H} A_l \bigl[\mathbin{\cdot} \mathbin{,} h \mathbin{,} -1 \mathbin{,} IDX^I \bigr],
\end{equation}
where \( IDX^I \) represents the indices of image tokens within the input sequence. To address the variability in the number of image tokens and the discrepancies in the overall attention weights assigned to the image across different models, we select the top \(\lambda\) proportion of image tokens with the highest attention weights as the model's most attended ones, with their indices denoted as \( IDX^I_{att} \):
\begin{equation}
IDX^I_{att} = \operatorname{argtopk}_{i \in IDX^I} ( A^I \bigl[ \mathbin{\cdot} \mathbin{,} i \bigr] \mathbin{,} \lambda),
\label{eq:high attn select}
\end{equation}
Given the original image tokens \(v\) and indices of the model's attended image tokens \( IDX^I_{att} \), we zero out the remaining image tokens, whose indices are \(IDX^I \setminus IDX^I_{att} \), resulting in the model's attended image tokens \(v_{att}\).


\subsection{Mixture of Decoding}

The decoding process is conducted within the LLM backbone of the LVLM, parameterized by \(\theta\). Given the image tokens \(v\), text input tokens \(x\), and previous output tokens \(y_{<t}\), the probability distribution for the \(t\)-th token \(y_t\) is expressed as:
\begin{equation}
    y_t \sim p_\theta(y_t \mid v, x, y_{<t}) \propto \exp \text{logit}_\theta(y_t \mid v, x, y_{<t}).
\end{equation}
Similarly, the probability distribution for \(y_t\) can also be derived based on the model's attended image tokens \(v_{att}\), text input tokens \(x\), and previous output tokens \(y_{<t}\), denoted as \(p_\theta(y_t \mid v_{att}, x, y_{<t})\). Subsequently, we define \(d(v, v_{att})\) as the JS divergence between the output probability distributions of \(y_t\) obtained from the two different sets of image tokens \(v\) and \(v_{att}\):
\begin{equation}
  \begin{aligned}
    d(v, v_{att}) = D_{JS} \bigl[ &p_\theta(y_t \mid v, x, y_{<t}) \parallel \\
    &p_\theta(y_t \mid v_{att}, x, y_{<t}) \bigr].
  \end{aligned}
\end{equation}
If the output based on \(v_{att}\) aligns with the original output derived from \(v\), it indicates that the model's attention over image tokens is right, as illustrated in the upper part of Fig.~\ref{fig:overview}. In this scenario, the two outputs are complemented to enhance the output probability of proper tokens. Conversely, if the outputs are inconsistent, it suggests that the model's attention over image tokens is wrong, as depicted in the lower part of Fig.~\ref{fig:overview}. In this case, the two outputs are contrasted to counteract the hallucination caused by the erroneous attention. Thus, the MoD strategy is formulated as follows:
\begin{equation}
  \begin{aligned}
    &y_t \sim p_\theta(y_t |~ v, v_{att}, x, y_{<t}) \\
      &=\begin{cases}
           \text{softmax} \bigl[ \text{logit}_\theta(y_t |~ v, x, y_{<t}) + \\ \alpha_1 \cdot \text{logit}_\theta(y_t |~ v_{att}, x, y_{<t}) \bigr], 
           \text{if~} d(v, v_{att}) \leq \gamma; \\
           \text{softmax} \bigl[ (1 + \alpha_2) \cdot \text{logit}_\theta(y_t |~ v, x, y_{<t}) - \\ 
            \alpha_2 \cdot \text{logit}_\theta(y_t |~ v_{att}, x, y_{<t}) \bigr],  \text{if~} d(v, v_{att}) > \gamma.
         \end{cases}
    \end{aligned}
  \label{eq:MoD}
\end{equation}
Here, \(\alpha_1\) and \(\alpha_2\) denote the hyperparameters for the complement and contrast operations, respectively, and \(\gamma\) represents the JS divergence threshold, which is used to determine the consistency between the two outputs. 

\section{Experiments}

\subsection{Experimental Settings}

\noindent\textbf{Benchmarks and metrics.}
(1)~\textbf{POPE}~\cite{li2023evaluating} is a widely used benchmark designed to evaluate object hallucination. It employs yes/no questions formulated as ``Is there an \{object\} in the image?'' to query the model about the presence of specific objects in an image. The evaluation metrics include Accuracy, Precision, Recall, and F1 score. 
(2)~\textbf{MME}~\cite{fu2023mme} is a more challenging benchmark for hallucination assessment. It designs a pair of similar questions with answers ``yes'' and ``no'', respectively, for each image. In addition to question-level accuracy (Acc), Acc+ is introduced to represent image-level accuracy. An image is deemed to be fully understood only if the model answers both associated questions correctly. The overall performance is quantified by the MME Score, which is calculated as the sum of Acc and Acc+. 
(3)~\textbf{CHAIR}~\cite{rohrbach2018object} evaluates object hallucination in image captioning tasks. In this evaluation framework, LVLMs are asked to generate descriptive captions for a randomly selected subset of 500 images from the MS-COCO validation set. Specifically, CHAIR quantifies hallucination by calculating the proportion of objects mentioned in the model's caption that do not exist in the ground truth. The main metrics comprise \(\text{CHAIR}_i\) and \(\text{CHAIR}_s\), which assess hallucination at the instance level and the sentence level, respectively.
(4)~\textbf{AMBER}~\cite{wang2023amber} is a more comprehensive benchmark that evaluates object hallucination, attribute hallucination, and relation hallucination in both discriminative and generative tasks. For discriminative tasks, evaluation metrics include Accuracy, Precision, Recall, and F1 score. For generative tasks, the metrics include \(\text{CHAIR}_i\), Cover, Hal, and Cog (see Appendix~\ref{sec: metric details} for details). Additionally, the AMBER Score is proposed to provide a unified evaluation, calculated as:
\begin{equation}
\text{AMBER Score} = \frac{1}{2} \times \left(1 - \text{CHAIR}_i + \text{F1}\right).
\end{equation}

\noindent\textbf{Models.}
The performance of MoD is evaluated on the 7B versions of LLaVA-1.5~\cite{liu2024llava}, Qwen-VL~\cite{bai2023qwenvl}, and LLaVA-NEXT~\cite{liu2024llavanext}, respectively. LLaVA-1.5 employs a two-layer MLP to align vision and language modalities, while Qwen-VL uses a single-layer cross-attention module to compress and align visual features to the language modality. LLaVA-NEXT further enhances LLaVA-1.5 by introducing the ``anyres'' mechanism, which increases the input image resolution by \(4\times\) to enable fine-grained visual understanding. Notably, MoD is model-agnostic and can be integrated into diverse LVLM architectures.

\noindent\textbf{Baselines.}
We first adopt sampling decoding as the most basic baseline, where the next token is sampled directly from the output probability distribution. Additionally, we compare our method with three contrastive decoding strategies: VCD~\cite{leng2024mitigating}, M3ID~\cite{favero2024multi}, and AvisC~\cite{woo2024don}. To ensure a fair comparison, we report the results of these methods based on our reproduction using their official codebases within our own experimental environment.

\begin{table*}[ht]
    \centering
    \resizebox{\linewidth}{!}{
    \begin{tabular}{l l @{\hspace{25pt}} c c c c @{\hspace{25pt}} c c c c @{\hspace{25pt}} c c c c}
    \toprule
        \multirow{2}{*}{\textbf{Setting}} & \multirow{2}{*}{\textbf{Method}} & \multicolumn{4}{c@{\hspace{25pt}}}{\textbf{LLaVA-v1.5}} & \multicolumn{4}{c@{\hspace{25pt}}}{\textbf{Qwen-VL}} & \multicolumn{4}{c@{\hspace{15pt}}}{\textbf{LLaVA-NEXT}} \\
        ~ & ~ & Acc & Pre & Rec & F1 & Acc & Pre & Rec & F1 & Acc & Pre & Rec & F1 \\
        \midrule
        \multirow{5}{*}{random} & sampling & 83.8 & 82.4 & 86.1 & 84.2 & 84.9 & 96.0 & 72.9 & 82.9 & 84.4 & 94.7 & 72.8 & 82.3 \\
        ~ & VCD & 85.0 & 82.7 & 86.1 & 84.2 & 85.5 & 96.0 & 71.1 & 83.6 & 86.0 & 96.5 & 74.8 & 84.3 \\
        ~ & M3ID & 86.1 & 83.2 & 86.8 & 85.0 & 85.3 & 95.1 & 74.2 & 83.4 & 85.5 & 96.3 & 73.9 & 83.6 \\
        ~ & AvisC & 82.3 & 78.4 & 89.3 & 83.5 & 82.9 & 96.2 & 68.5 & 80.0 & 85.2 & 98.6 & 71.3 & 82.8 \\
        ~ & MoD (Ours) & \textbf{89.2} & 90.0 & 88.2 & \textbf{89.1} & \textbf{86.0} & 97.0 & 74.3 & \textbf{84.1} & \textbf{86.6} & 97.7 & 74.9 & \textbf{84.8} \\
        \midrule
        \multirow{5}{*}{popular} & sampling & 82.0 & 79.7 & 85.9 & 82.6 & 84.0 & 94.7 & 72.1 & 81.9 & 83.2 & 90.9 & 73.8 & 81.5 \\
        ~ & VCD & 82.1 & 78.5 & 88.3 & 83.2 & 84.9 & 94.5 & 74.9 & 83.6 & 84.5 & 92.9 & 74.8 & 82.9 \\
        ~ & M3ID & 82.8 & 80.1 & 88.5 & 84.1 & 84.2 & 94.1 & 73.8 & 82.7 & 84.2 & 93.1 & 73.9 & 82.4 \\
        ~ & AvisC & 78.2 & 72.7 & 90.3 & 80.5 & 82.8 & 95.5 & 68.9 & 80.1 & 83.9 & 94.7 & 71.7 & 81.6 \\
        ~ & MoD (Ours) & \textbf{85.7} & 84.1 & 88.1 & \textbf{86.1} & \textbf{85.6} & 96.3 & 74.0 & \textbf{83.7} & \textbf{85.5} & 95.1 & 74.9 & \textbf{83.8} \\
        \midrule
        \multirow{5}{*}{adversarial} & sampling & 75.8 & 71.3 & 86.3 & 78.1 & 82.1 & 90.0 & 72.3 & 80.2 & 79.5 & 84.1 & 72.9 & 78.1 \\
        ~ & VCD & 76.3 & 71.5 & 87.3 & 78.7 & 84.0 & 90.6 & 74.9 & 82.0 & 80.9 & 85.2 & 74.8 & 79.7 \\
        ~ & M3ID & 77.1 & 71.8 & 87.6 & 78.9 & 83.2 & 90.4 & 73.1 & 80.8 & 80.6 & 85.4 & 73.9 & 79.2 \\
        ~ & AvisC & 74.2 & 68.4 & 89.9 & 77.7 & 81.2 & 91.9 & 68.5 & 78.5 & 81.8 & 91.3 & 71.5 & 80.2 \\
        ~ & MoD (Ours) & \textbf{79.7} & 75.4 & 88.2 & \textbf{81.3} & \textbf{84.0} & 92.4 & 74.2 & \textbf{82.3} & \textbf{82.4} & 88.3 & 74.8 & \textbf{81.0} \\
    \bottomrule
    \end{tabular}
    }
    \caption{\textbf{Results on POPE benchmark.} ``Acc'', ``Pre'', ``Rec'', and ``F1'' stand for Accuracy, Precision, Recall, and F1 score, respectively. The reported results are derived from the MS-COCO dataset.}
    \label{tab:POPE result}
\end{table*}

\noindent\textbf{Implementation Details.}
In contrast to many existing decoding methods that necessitate extensive task- and model-specific hyperparameter tuning, MoD demonstrates robust performance across diverse tasks and models using a shared set of hyperparameters, with low sensitivity to their selection (see Sec.~\ref{sec: ablation} for detailed analysis). Specifically, for the extraction of the model's attended image tokens (Eq.~\ref{eq:high attn select}), we set \(\lambda = 0.2\); for the decoding process (Eq.~\ref{eq:MoD}), we set \(\alpha_1 = 4\), \(\alpha_2 = 1\), and \(\gamma = 0.05\). All experiments are conducted on NVIDIA RTX 3090 GPUs with 24GB of memory, utilizing a fixed random seed of 42 for reproducibility.

\subsection{Experimental Results}

\noindent\textbf{Results on POPE.}
Tab.~\ref{tab:POPE result} presents the performance of various decoding methods on the POPE benchmark based on the MS-COCO dataset. The complete experimental results are provided in Appendix~\ref{sec:detailed pope}. As shown, MoD achieves the best performance across all three evaluation settings, outperforming the second-best method by up to 3.1 points in Accuracy and 4.1 points in F1 score. These results demonstrate that MoD enhances the model's ability to perceive fine-grained object details in images, effectively mitigating object hallucinations. Additionally, MoD exhibits remarkable superiority in Precision, surpassing other methods by up to 6.8 points, indicating fewer false positive samples. This suggests that the model becomes more conservative when generating positive answers, effectively suppressing the tendency of existing LVLMs to answer ``Yes'' without adequately considering the question itself.

\begin{table}
    \centering
    \resizebox{\columnwidth}{!}{
    \begin{tabular}{l l c c c c c}
    \toprule
        \multirow{2}{*}{\textbf{Model}} & \multirow{2}{*}{\textbf{Method}} & \multicolumn{2}{c}{\textbf{Object-level}} & \multicolumn{2}{c}{\textbf{Attribute-level}} & \multirow{2}{*}{\textbf{Total}} \\ 
        ~ & ~ & Existence & Count & Position & Color & ~ \\
        \midrule
        \multirow{5}{*}{\rotatebox{90}{LLaVA-v1.5}} & sampling & 170.0 & 103.3 & 108.3 & 128.3 & 510.0 \\
        ~ & VCD & 180.0 & 110.0 & 108.3 & 133.3 & 531.7 \\
        ~ & M3ID & 185.0 & 118.3 & 121.7 & 128.3 & 553.3 \\
        ~ & AvisC & 195.0 & 116.7 & 131.7 & 153.3 & 596.7 \\
        ~ & MoD (Ours) & 195.0 & 141.7 & 126.7 & 175.0 & \textbf{638.3} \\
        \midrule
        \multirow{5}{*}{\rotatebox{90}{Qwen-VL}} & sampling & 160.0 & 143.3 & 113.3 & 165.0 & 581.7 \\
        ~ & VCD & 165.0 & 140.0 & 113.3 & 175.0 & 593.3 \\
        ~ & M3ID & 165.0 & 143.3 & 103.3 & 175.0 & 586.7 \\
        ~ & AvisC & 160.0 & 145.0 & 113.3 & 160.0 & 578.3 \\
        ~ & MoD (Ours) & 170.0 & 160.0 & 103.3 & 180.0 & \textbf{613.3} \\
        \midrule
        \multirow{5}{*}{\rotatebox{90}{LLaVA-NEXT}} & sampling & 175.0 & 143.3 & 131.7 & 145.0 & 595.0 \\
        ~ & VCD & 190.0 & 145.0 & 116.7 & 160.0 & 611.7 \\
        ~ & M3ID & 195.0 & 145.0 & 103.3 & 165.0 & 608.3 \\
        ~ & AvisC & 195.0 & 160.0 & 108.3 & 150.0 & 613.3 \\
        ~ & MoD (Ours) & 195.0 & 160.0 & 133.3 & 165.0 & \textbf{653.3} \\
    \bottomrule
    \end{tabular}
    }
    \caption{\textbf{Results on MME benchmark.} The performance is measured by MME Score. The ``Total'' column represents the sum of four individual results in each row.}
    \label{tab:MME result}
\end{table}

\noindent\textbf{Results on MME.}
Following prior studies, we evaluate object and attribute hallucination on four subsets of the MME benchmark, as detailed in Tab.~\ref{tab:MME result}. Overall, MoD shows significant improvements over the second-best method, with notable gains in MME Score of 41.6 points for LLaVA-1.5, 20.0 points for Qwen-VL, and 40.0 points for LLaVA-NEXT. However, there is a slight decline in the ``position'' subset for LLaVA-1.5 and Qwen-VL. We hypothesize that this is due to the masking of low-attention image tokens, which employs a zeroing-out approach, potentially reducing the model's sensitivity to relative positional information. This issue might be addressed by refining the masking strategy, such as incorporating pooling mechanisms to allow the model to retain partial awareness of low-attention image tokens. We leave this exploration for future work. Notably, this decline is not observed in LLaVA-NEXT, which we attribute to its ``anyres'' mechanism, where the masked local and global image tokens complement each other, thereby preserving relative positional information. Despite this, MoD consistently outperforms other decoding methods on the remaining three subsets and achieves the highest total MME Score, demonstrating its overall effectiveness in mitigating hallucinations.

\begin{figure}
  \includegraphics[width=\linewidth]{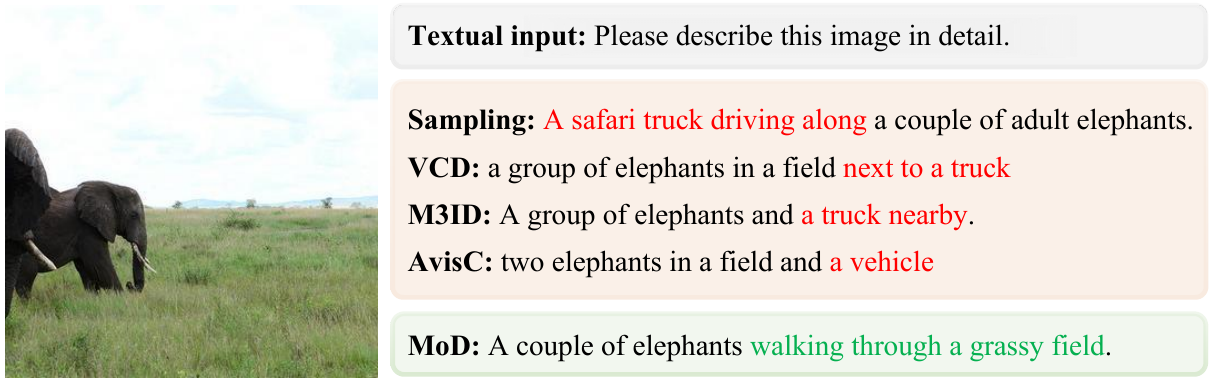}
  \caption {\textbf{Case study of generative tasks using Qwen-VL.} We compare responses generated by sampling, VCD, M3ID, AvisC, and our proposed MoD. Hallucinated content is highlighted in 
\textcolor[HTML]{ff0000}{red}, while more detailed and accurate content is marked in 
\textcolor[HTML]{06b050}{green}.}
  \label{fig:casestudy2}
\end{figure}

\noindent\textbf{Results on CHAIR.}
In addition to its strong performance in discriminative tasks, MoD also exhibits exceptional capabilities in generative tasks. As evidenced by the quantitative results in Tab.~\ref{tab:CHAIR result}, MoD achieves the best performance across both sentence-level (\(\text{CHAIR}_s\)) and instance-level (\(\text{CHAIR}_i\)) hallucination evaluation metrics. Moreover, MoD also excels in Recall, which measures the completeness of generated captions. Notably, despite using the prompt ``Please describe this image in detail'', Qwen-VL tends to produce concise outputs, resulting in significantly lower absolute values for all metrics. However, MoD still manages to reduce hallucination in Qwen-VL while simultaneously improving its output completeness. Qualitative results presented in Fig.~\ref{fig:casestudy2} further illustrate this point, demonstrating the effectiveness of MoD regardless of the caption length. Additionally, while M3ID intensifies hallucinations in LLaVA-1.5, and both VCD and AvisC exacerbate hallucinations in LLaVA-NEXT, MoD consistently mitigates hallucinations across all models, highlighting its robustness and model-agnostic characteristics.

\begin{table}
    \centering
    \resizebox{\columnwidth}{!}{
    \begin{tabular}{l l c c c c}
    \toprule
        \textbf{Model} & \textbf{Method} & CHAIR$_s\downarrow$ & CHAIR$_i\downarrow$ & Recall~$\uparrow$ & Length \\
        \midrule
        \multirow{5}{*}{\rotatebox{90}{LLaVA-v1.5}} & sampling & 52.8 & 15.9 & 77.3 & 93.4 \\
        ~ & VCD & 51.0 & 14.9 & 77.2 & 101.9 \\
        ~ & M3ID & 56.2 & 17.0 & \textbf{79.3} & 97.1  \\
        ~ & AvisC & 44.0 & 13.7 & 72.9 & 89.8 \\
        ~ & MoD (Ours) & \textbf{42.6} & \textbf{12.4} & 78.9 & 97.6 \\
        \midrule
        \multirow{5}{*}{\rotatebox{90}{Qwen-VL}} & sampling & 2.8 & 3.0 & 31.0 & 5.3 \\
        ~ & VCD & 1.4 & 1.2 & 30.8 & 4.0 \\
        ~ & M3ID & 1.7 & 1.3 & 31.8 & 3.4 \\
        ~ & AvisC & 1.6 & 1.6 & 32.0 & 4.4 \\
        ~ & MoD (Ours) & \textbf{0.8} & \textbf{1.0} & \textbf{32.1} & 3.8 \\
        \midrule
        \multirow{5}{*}{\rotatebox{90}{LLaVA-NEXT}} & sampling & 35.8 & 12.0 & 59.5 & 179.0 \\
        ~ & VCD & 40.2 & 10.7 & \textbf{62.1} & 171.2 \\
        ~ & M3ID & 35.2 & 10.3 & 61.4 & 152.7 \\
        ~ & AvisC & 40.4 & 12.4 & 60.0 & 183.8 \\
        ~ & MoD (Ours) & \textbf{33.6} & \textbf{9.6} & 61.4 & 174.9 \\
    \bottomrule
    \end{tabular}
    }
    \caption{\textbf{Results on CHAIR benchmark.} Lower \(\text{CHAIR}_s\) and \(\text{CHAIR}_i\), along with higher Recall, correspond to better performance.}
    \label{tab:CHAIR result}
\end{table}

\begin{table*}[ht]
    \centering
    \resizebox{\linewidth}{!}{
    \begin{tabular}{l l @{\hspace{20pt}} c c c c @{\hspace{20pt}} c c c c @{\hspace{20pt}} c}
    \toprule
        \multirow{2}{*}{\textbf{Model}} & \multirow{2}{*}{\textbf{Method}} & \multicolumn{4}{c@{\hspace{20pt}}}{\textbf{Discriminative}} & \multicolumn{4}{c}{\textbf{Generative}} & \multirow{2}{*}{\makecell{\textbf{AMBER} \\ \textbf{Score}}}\\ 
        ~ & ~ & Acc & Pre & Rec & F1 & CHAIR$_i\downarrow$ & Cover~$\uparrow$ & Hal~$\downarrow$ & Cog* \\
        \midrule
        \multirow{5}{*}{LLaVA-v1.5} & sampling & 67.0 & 85.2 & 60.9 & 71.0 & 12.0 & 50.3 & 51.0 & 4.6 & 79.5 \\
        ~ & VCD & 67.3 & 86.1 & 60.5 & 71.1 & 10.0 & \textbf{51.2} & 43.6 & 4.3 & 80.6 \\
        ~ & M3ID & 67.3 & 86.5 & 60.1 & 70.9 & 9.1 & 49.4 & 42.8 & 4.4 & 80.9 \\
        ~ & AvisC & 70.7 & 85.5 & 65.6 & 74.2 & 11.8 & 50.4 & 51.2 & 5.0 & 81.2 \\
        ~ & MoD (Ours) & \textbf{72.4} & 91.5 & 64.0 & \textbf{75.3} & \textbf{8.6} & 50.7 & \textbf{38.8} & 4.7 & \textbf{83.4} \\
        \midrule
        \multirow{5}{*}{Qwen-VL} & sampling & 82.9 & 88.0 & 85.9 & 86.9 & 4.8 & 31.3 & 9.2 & 0.3 & 91.1 \\
        ~ & VCD & 84.1 & 89.2 & 86.6 & 87.9 & 3.5 & 35.2 & 7.7 & 0.3 & 92.2 \\
        ~ & M3ID & 84.0 & 88.1 & 86.2 & 87.1 & 3.7 & 34.1 & 8.2 & 0.4 & 91.7 \\
        ~ & AvisC & 84.1 & 88.8 & 87.0 & 87.9 & 4.3 & 35.0 & 8.6 & 0.4 & 91.8 \\
        ~ & MoD (Ours) & \textbf{84.2} & 89.2 & 86.7 & \textbf{87.9} & \textbf{2.2} & \textbf{36.5} & \textbf{4.5} & 0.3 & \textbf{92.9} \\
        \midrule
        \multirow{5}{*}{LLaVA-NEXT} & sampling & 72.9 & 82.4 & 75.2 & 78.6 & 12.0 & 56.5 & 59.6 & 5.1 & 83.3 \\
        ~ & VCD & 74.3 & 83.9 & 75.8 & 79.6 & 11.8 & 59.1 & 58.6 & 5.0 & 83.9 \\
        ~ & M3ID & 75.0 & 84.9 & 75.7 & 80.0 & 10.2 & 59.0 & 52.8 & 4.6 & 84.9 \\
        ~ & AvisC & 77.7 & 88.0 & 76.9 & 82.1 & 14.2 & 58.1 & 65.2 & 5.7 & 84.0 \\
        ~ & MoD (Ours) & \textbf{78.9} & 83.4 & 85.2 & \textbf{84.3} & \textbf{9.4} & \textbf{61.6} & \textbf{50.8} & 3.2 & \textbf{87.5} \\
    \bottomrule
    \end{tabular}
    }
    \caption{\textbf{Results on AMBER benchmark.} In discriminative tasks, ``Acc'', ``Pre'', ``Rec'', and ``F1'' stand for Accuracy, Precision, Recall, and F1 score, respectively. Higher values for these metrics indicate superior performance. In generative tasks, lower \(\text{CHAIR}_i\) and Hal, along with higher Cover, signify better performance. *Cog measures the extent to which hallucinations in LVLMs align with human cognition and is therefore not directly comparable.}
    \label{tab:AMBER result}
\end{table*}

\noindent\textbf{Results on AMBER.}
The AMBER benchmark evaluates both discriminative and generative tasks, with quantitative results shown in Tab.~\ref{tab:AMBER result}. In discriminative tasks, MoD maintains the leading performance in Accuracy and F1 score, achieving average improvements of 4.2 and 3.7 points, respectively, across the three evaluated models. In generative tasks, MoD significantly outperforms other methods in the two hallucination evaluation metrics, \(\text{CHAIR}_i\) and Hal, with notable average performance gains of 2.9 and 2.1 points, respectively, while maintaining excellent coverage. Overall, MoD attains the highest AMBER score across all evaluated models, with improvements of 2.2 points for LLaVA-1.5, 0.7 points for Qwen-VL, and 2.6 points for LLaVA-NEXT compared to the second-best method, showcasing its versatility in handling both discriminative and generative tasks. Additional qualitative results, provided in Appendix~\ref{sec: more cases}, further illustrate that MoD not only effectively mitigates hallucinations but also enriches the detail and depth of generated image captions.

\subsection{Ablation Studies}
\label{sec: ablation}

\begin{table}[ht]
    \centering
    \resizebox{\columnwidth}{!}{
    \begin{tabular}{l l c c c c c}
    \toprule
        \multirow{2}{*}{\textbf{Model}} & \multirow{2}{*}{\textbf{Method}} & \multicolumn{2}{c}{\textbf{Object-level}} & \multicolumn{2}{c}{\textbf{Attribute-level}} & \multirow{2}{*}{\textbf{Total}} \\ 
        ~ & ~ & Existence & Count & Position & Color & ~ \\
        \midrule
        \multirow{4}{*}{\rotatebox{90}{\makecell{LLaVA\\-v1.5}}} & sampling & 170.0 & 103.3 & 108.3 & 128.3 & 510.0 \\
        ~ & complementary & 195.0 & 143.3 & 116.7 & 160.0 & 615.0 \\
        ~ & contrastive & 190.0 & 145.0 & 120.0 & 165.0 & 620.0 \\
        ~ & MoD (Ours) & 195.0 & 141.7 & 126.7 & 175.0 & \textbf{638.3} \\
        \midrule
        \multirow{4}{*}{\rotatebox{90}{\makecell{Qwen\\-VL}}} & sampling & 160.0 & 143.3 & 113.3 & 165.0 & 581.7 \\
        ~ & complementary & 175.0 & 145.0 & 103.3 & 180.0 & 603.3 \\
        ~ & contrastive & 175.0 & 155.0 & 103.3 & 170.0 & 603.3 \\
        ~ & MoD (Ours) & 170.0 & 160.0 & 103.3 & 180.0 & \textbf{613.3} \\
        \midrule
        \multirow{4}{*}{\rotatebox{90}{\makecell{LLaVA\\-NEXT}}} & sampling & 175.0 & 143.3 & 131.7 & 145.0 & 595.0 \\
        ~ & complementary & 195.0 & 145.0 & 133.3 & 165.0 & 638.3 \\
        ~ & contrastive & 190.0 & 150.0 & 138.3 & 155.0 & 633.3 \\
        ~ & MoD (Ours) & 195.0 & 160.0 & 133.3 & 165.0 & \textbf{653.3} \\
    \bottomrule
    \end{tabular}
    }
    \caption{\textbf{Ablation study of individual decoding methods.} The performance of sampling, complementary, and contrastive decoding methods are compared with MoD on MME benchmark across three LVLMs.}
    \label{tab:PH&MH ablation}
\end{table}

\noindent\textbf{Comparison with individual methods.}
To evaluate the performance gains brought by MoD compared to employing either the complementary method or the contrastive method independently, which corresponds to setting the consistency threshold \(\gamma\) in Eq.~\ref{eq:MoD} to 1 and 0, respectively, we conduct ablation studies on the MME benchmark. The results, as detailed in Tab~\ref{tab:PH&MH ablation}, demonstrate that MoD effectively integrates the strengths of both individual methods, achieving maximum performance improvements of 23.3 points over the complementary method alone and 20.0 points over the contrastive method alone. This underscores the effectiveness of MoD in implementing adaptive decoding strategies by assessing the correctness of the model's attention over image tokens.

\begin{figure}[h]
  \includegraphics[width=\columnwidth]{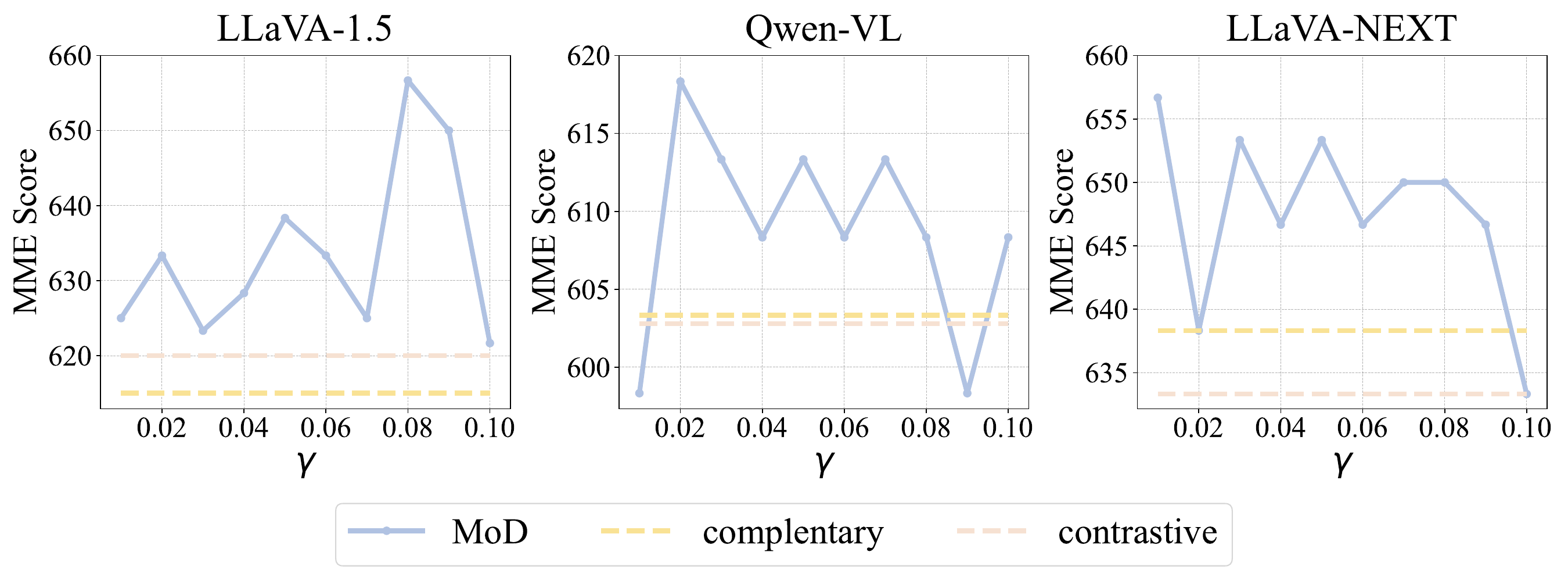}
  \caption{\textbf{Ablation study on the consistency threshold \(\gamma\).} We vary \(\gamma\) from 0.01 to 0.10 with an increment of 0.01, and compare the results with those obtained using individual methods on MME benchmark. In MoD, \(\gamma\) is consistently set to 0.05 across all experiments.}
  \label{fig:gamma ablation}
\end{figure}

\noindent\textbf{Analysis of the consistency threshold \(\gamma\).}
Fig.~\ref{fig:gamma ablation} presents a comprehensive evaluation of the performance of MoD across varying consistency thresholds on the MME benchmark. The experimental results indicate that MoD maintains a consistent performance advantage over a wide range of consistency thresholds from 0.02 to 0.08, significantly outperforming both complementary and contrastive methods when employed independently. Specifically, the average MME Score achieved by MoD exceeds that of the best individual method by 14.0 points for LLaVA-1.5, 8.6 points for Qwen-VL, and 10.0 points for LLaVA-NEXT, underscoring the robustness of MoD across varying consistency threshold configurations.

We provide a detailed analysis of additional hyperparameters of MoD in Appendix~\ref{sec: other ablas}. Notably, MoD employs a shared set of hyperparameters across different tasks and model architectures, distinguishing it from most other decoding methods that necessitate meticulous hyperparameter tuning for specific scenarios. Furthermore, experimental results demonstrate that MoD exhibits low sensitivity to hyperparameter variations, further highlighting its practicality and adaptability.

\section{Conclusion}
In this paper, we propose Mixture of Decoding (MoD), an adaptive decoding strategy designed to mitigate hallucinations in LVLMs. By evaluating the consistency between the outputs generated from the original image tokens and those derived from the model's attended image tokens, MoD initially assesses the correctness of the model's attention over image tokens. Subsequently, informed by this evaluation, MoD dynamically selects either the complementary or contrastive decoding strategy to optimize the decoding process. A series of comprehensive experiments have been conducted to validate the effectiveness and robustness of MoD, highlighting its potential as a promising solution to enhance the accuracy and reliability of LVLMs across diverse applications.

\section*{Limitations}
While MoD demonstrates promising results, we acknowledge several limitations of our approach.

First, similar to other decoding methods, MoD requires two forward passes, nearly doubling the computational latency compared to standard decoding. However, MoD achieves significantly better performance under the same inference cost.

Additionally, our current implementation employs a straightforward masking strategy that zeros out low-attention image tokens. Exploring more sophisticated masking strategies, such as pooling mechanisms, could potentially further enhance performance across a broader range of tasks. We leave this investigation for future work.

\section*{Ethical Considerations}
Although MoD is designed to mitigate hallucinations and improve the reliability of LVLMs, it does not inherently address biases embedded in training data, which may lead to unfair or discriminatory outcomes. We emphasize the critical importance of prioritizing fairness and responsible deployment to ensure that LVLMs, augmented by MoD, contribute positively to society while minimizing potential ethical risks.

\section*{Acknowledgements}
This work is jointly sponsored by National Natural Science Foundation of China (62236010, 62141608, 62206291).

\bibliography{custom}

\begin{thebibliography}{41}
\providecommand{\natexlab}[1]{#1}

\bibitem[{Bai et~al.(2023{\natexlab{a}})Bai, Bai, Chu, Cui, Dang, Deng, Fan, Ge, Han, Huang et~al.}]{bai2023qwen}
Jinze Bai, Shuai Bai, Yunfei Chu, Zeyu Cui, Kai Dang, Xiaodong Deng, Yang Fan, Wenbin Ge, Yu~Han, Fei Huang, et~al. 2023{\natexlab{a}}.
\newblock Qwen technical report.
\newblock \emph{arXiv preprint arXiv:2309.16609}.

\bibitem[{Bai et~al.(2023{\natexlab{b}})Bai, Bai, Yang, Wang, Tan, Wang, Lin, Zhou, and Zhou}]{bai2023qwenvl}
Jinze Bai, Shuai Bai, Shusheng Yang, Shijie Wang, Sinan Tan, Peng Wang, Junyang Lin, Chang Zhou, and Jingren Zhou. 2023{\natexlab{b}}.
\newblock Qwen-vl: A frontier large vision-language model with versatile abilities.
\newblock \emph{arXiv preprint arXiv:2308.12966}.

\bibitem[{Bai et~al.(2024)Bai, Wang, Xiao, He, Han, Zhang, and Shou}]{bai2024hallucination}
Zechen Bai, Pichao Wang, Tianjun Xiao, Tong He, Zongbo Han, Zheng Zhang, and Mike~Zheng Shou. 2024.
\newblock Hallucination of multimodal large language models: A survey.
\newblock \emph{arXiv preprint arXiv:2404.18930}.

\bibitem[{Chen et~al.(2024{\natexlab{a}})Chen, Li, Zhang, Liu, Li, Gao, Hong, Tian, Zhao, Li et~al.}]{chen2024automated}
Kai Chen, Yanze Li, Wenhua Zhang, Yanxin Liu, Pengxiang Li, Ruiyuan Gao, Lanqing Hong, Meng Tian, Xinhai Zhao, Zhenguo Li, et~al. 2024{\natexlab{a}}.
\newblock Automated evaluation of large vision-language models on self-driving corner cases.
\newblock \emph{arXiv preprint arXiv:2404.10595}.

\bibitem[{Chen et~al.(2024{\natexlab{b}})Chen, Ma, Zhang, Xu, Qian, Yang, Fouhey, and Chai}]{chen2024multi}
Xuweiyi Chen, Ziqiao Ma, Xuejun Zhang, Sihan Xu, Shengyi Qian, Jianing Yang, David~F Fouhey, and Joyce Chai. 2024{\natexlab{b}}.
\newblock Multi-object hallucination in vision-language models.
\newblock \emph{arXiv preprint arXiv:2407.06192}.

\bibitem[{Chuang et~al.(2023)Chuang, Xie, Luo, Kim, Glass, and He}]{chuang2023dola}
Yung-Sung Chuang, Yujia Xie, Hongyin Luo, Yoon Kim, James Glass, and Pengcheng He. 2023.
\newblock Dola: Decoding by contrasting layers improves factuality in large language models.
\newblock \emph{arXiv preprint arXiv:2309.03883}.

\bibitem[{Dhuliawala et~al.(2023)Dhuliawala, Komeili, Xu, Raileanu, Li, Celikyilmaz, and Weston}]{dhuliawala2023chain}
Shehzaad Dhuliawala, Mojtaba Komeili, Jing Xu, Roberta Raileanu, Xian Li, Asli Celikyilmaz, and Jason Weston. 2023.
\newblock Chain-of-verification reduces hallucination in large language models.
\newblock \emph{arXiv preprint arXiv:2309.11495}.

\bibitem[{Favero et~al.(2024)Favero, Zancato, Trager, Choudhary, Perera, Achille, Swaminathan, and Soatto}]{favero2024multi}
Alessandro Favero, Luca Zancato, Matthew Trager, Siddharth Choudhary, Pramuditha Perera, Alessandro Achille, Ashwin Swaminathan, and Stefano Soatto. 2024.
\newblock Multi-modal hallucination control by visual information grounding.
\newblock In \emph{Proceedings of the IEEE/CVF Conference on Computer Vision and Pattern Recognition}, pages 14303--14312.

\bibitem[{Fu et~al.(2023)Fu, Chen, Shen, Qin, Zhang, Lin, Yang, Zheng, Li, Sun et~al.}]{fu2023mme}
Chaoyou Fu, Peixian Chen, Yunhang Shen, Yulei Qin, Mengdan Zhang, Xu~Lin, Jinrui Yang, Xiawu Zheng, Ke~Li, Xing Sun, et~al. 2023.
\newblock Mme: A comprehensive evaluation benchmark for multimodal large language models.
\newblock \emph{arXiv preprint arXiv:2306.13394}.

\bibitem[{He et~al.(2023)He, Mao, Lin, Ruan, Lan, Feng, and Cambria}]{he2023survey}
Kai He, Rui Mao, Qika Lin, Yucheng Ruan, Xiang Lan, Mengling Feng, and Erik Cambria. 2023.
\newblock A survey of large language models for healthcare: from data, technology, and applications to accountability and ethics.
\newblock \emph{arXiv preprint arXiv:2310.05694}.

\bibitem[{Huang et~al.(2023)Huang, Yu, Ma, Zhong, Feng, Wang, Chen, Peng, Feng, Qin et~al.}]{huang2023survey}
Lei Huang, Weijiang Yu, Weitao Ma, Weihong Zhong, Zhangyin Feng, Haotian Wang, Qianglong Chen, Weihua Peng, Xiaocheng Feng, Bing Qin, et~al. 2023.
\newblock A survey on hallucination in large language models: Principles, taxonomy, challenges, and open questions.
\newblock \emph{ACM Transactions on Information Systems}.

\bibitem[{Huo et~al.(2024)Huo, Xu, Zhang, Wang, Chen, and Zhao}]{huo2024self}
Fushuo Huo, Wenchao Xu, Zhong Zhang, Haozhao Wang, Zhicheng Chen, and Peilin Zhao. 2024.
\newblock Self-introspective decoding: Alleviating hallucinations for large vision-language models.
\newblock \emph{arXiv preprint arXiv:2408.02032}.

\bibitem[{Jiang et~al.(2024)Jiang, Xu, Dong, Chen, Ye, Yan, Ye, Zhang, Huang, and Zhang}]{jiang2024hallucination}
Chaoya Jiang, Haiyang Xu, Mengfan Dong, Jiaxing Chen, Wei Ye, Ming Yan, Qinghao Ye, Ji~Zhang, Fei Huang, and Shikun Zhang. 2024.
\newblock Hallucination augmented contrastive learning for multimodal large language model.
\newblock In \emph{Proceedings of the IEEE/CVF Conference on Computer Vision and Pattern Recognition}, pages 27036--27046.

\bibitem[{Kim et~al.(2024)Kim, Kim, Lee, and Yun}]{kim2024instructive}
Taehyeon Kim, Joonkee Kim, Gihun Lee, and Se-Young Yun. 2024.
\newblock Instructive decoding: Instruction-tuned large language models are self-refiner from noisy instructions.
\newblock In \emph{The Twelfth International Conference on Learning Representations}.

\bibitem[{Leng et~al.(2024)Leng, Zhang, Chen, Li, Lu, Miao, and Bing}]{leng2024mitigating}
Sicong Leng, Hang Zhang, Guanzheng Chen, Xin Li, Shijian Lu, Chunyan Miao, and Lidong Bing. 2024.
\newblock Mitigating object hallucinations in large vision-language models through visual contrastive decoding.
\newblock In \emph{Proceedings of the IEEE/CVF Conference on Computer Vision and Pattern Recognition}, pages 13872--13882.

\bibitem[{Li et~al.(2024)Li, Patel, Vi{\'e}gas, Pfister, and Wattenberg}]{li2024inference}
Kenneth Li, Oam Patel, Fernanda Vi{\'e}gas, Hanspeter Pfister, and Martin Wattenberg. 2024.
\newblock Inference-time intervention: Eliciting truthful answers from a language model.
\newblock \emph{Advances in Neural Information Processing Systems}, 36.

\bibitem[{Li et~al.(2022)Li, Holtzman, Fried, Liang, Eisner, Hashimoto, Zettlemoyer, and Lewis}]{li2022contrastive}
Xiang~Lisa Li, Ari Holtzman, Daniel Fried, Percy Liang, Jason Eisner, Tatsunori Hashimoto, Luke Zettlemoyer, and Mike Lewis. 2022.
\newblock Contrastive decoding: Open-ended text generation as optimization.
\newblock \emph{arXiv preprint arXiv:2210.15097}.

\bibitem[{Li et~al.(2023)Li, Du, Zhou, Wang, Zhao, and Wen}]{li2023evaluating}
Yifan Li, Yifan Du, Kun Zhou, Jinpeng Wang, Wayne~Xin Zhao, and Ji-Rong Wen. 2023.
\newblock Evaluating object hallucination in large vision-language models.
\newblock \emph{arXiv preprint arXiv:2305.10355}.

\bibitem[{Liu et~al.(2024{\natexlab{a}})Liu, Li, Li, Li, Zhang, Shen, and Lee}]{liu2024llavanext}
Haotian Liu, Chunyuan Li, Yuheng Li, Bo~Li, Yuanhan Zhang, Sheng Shen, and Yong~Jae Lee. 2024{\natexlab{a}}.
\newblock \href {https://llava-vl.github.io/blog/2024-01-30-llava-next/} {Llava-next: Improved reasoning, ocr, and world knowledge}.

\bibitem[{Liu et~al.(2024{\natexlab{b}})Liu, Li, Wu, and Lee}]{liu2024llava}
Haotian Liu, Chunyuan Li, Qingyang Wu, and Yong~Jae Lee. 2024{\natexlab{b}}.
\newblock Visual instruction tuning.
\newblock \emph{Advances in neural information processing systems}, 36.

\bibitem[{Liu et~al.(2025)Liu, Chen, Ding, Xu, Wu, and Wang}]{liu2025attention}
Qiang Liu, Xinlong Chen, Yue Ding, Shizhen Xu, Shu Wu, and Liang Wang. 2025.
\newblock Attention-guided self-reflection for zero-shot hallucination detection in large language models.
\newblock \emph{arXiv preprint arXiv:2501.09997}.

\bibitem[{Luo et~al.(2023)Luo, Xiao, and Ma}]{luo2023zero}
Junyu Luo, Cao Xiao, and Fenglong Ma. 2023.
\newblock Zero-resource hallucination prevention for large language models.
\newblock \emph{arXiv preprint arXiv:2309.02654}.

\bibitem[{Manakul et~al.(2023)Manakul, Liusie, and Gales}]{manakul2023selfcheckgpt}
Potsawee Manakul, Adian Liusie, and Mark~JF Gales. 2023.
\newblock Selfcheckgpt: Zero-resource black-box hallucination detection for generative large language models.
\newblock \emph{arXiv preprint arXiv:2303.08896}.

\bibitem[{Rawte et~al.(2023)Rawte, Sheth, and Das}]{rawte2023survey}
Vipula Rawte, Amit Sheth, and Amitava Das. 2023.
\newblock A survey of hallucination in large foundation models.
\newblock \emph{arXiv preprint arXiv:2309.05922}.

\bibitem[{Rohrbach et~al.(2018)Rohrbach, Hendricks, Burns, Darrell, and Saenko}]{rohrbach2018object}
Anna Rohrbach, Lisa~Anne Hendricks, Kaylee Burns, Trevor Darrell, and Kate Saenko. 2018.
\newblock Object hallucination in image captioning.
\newblock \emph{arXiv preprint arXiv:1809.02156}.

\bibitem[{Sun et~al.(2023)Sun, Shen, Cao, Liu, Li, Shen, Gan, Gui, Wang, Yang et~al.}]{sun2023aligning}
Zhiqing Sun, Sheng Shen, Shengcao Cao, Haotian Liu, Chunyuan Li, Yikang Shen, Chuang Gan, Liang-Yan Gui, Yu-Xiong Wang, Yiming Yang, et~al. 2023.
\newblock Aligning large multimodal models with factually augmented rlhf.
\newblock \emph{arXiv preprint arXiv:2309.14525}.

\bibitem[{Tian et~al.(2024)Tian, Gu, Li, Liu, Wang, Zhao, Zhan, Jia, Lang, and Zhao}]{tian2024drivevlm}
Xiaoyu Tian, Junru Gu, Bailin Li, Yicheng Liu, Yang Wang, Zhiyong Zhao, Kun Zhan, Peng Jia, Xianpeng Lang, and Hang Zhao. 2024.
\newblock Drivevlm: The convergence of autonomous driving and large vision-language models.
\newblock \emph{arXiv preprint arXiv:2402.12289}.

\bibitem[{Touvron et~al.(2023)Touvron, Lavril, Izacard, Martinet, Lachaux, Lacroix, Rozi{\`e}re, Goyal, Hambro, Azhar et~al.}]{touvron2023llama}
Hugo Touvron, Thibaut Lavril, Gautier Izacard, Xavier Martinet, Marie-Anne Lachaux, Timoth{\'e}e Lacroix, Baptiste Rozi{\`e}re, Naman Goyal, Eric Hambro, Faisal Azhar, et~al. 2023.
\newblock Llama: Open and efficient foundation language models.
\newblock \emph{arXiv preprint arXiv:2302.13971}.

\bibitem[{Varshney et~al.(2023)Varshney, Yao, Zhang, Chen, and Yu}]{varshney2023stitch}
Neeraj Varshney, Wenlin Yao, Hongming Zhang, Jianshu Chen, and Dong Yu. 2023.
\newblock A stitch in time saves nine: Detecting and mitigating hallucinations of llms by validating low-confidence generation.
\newblock \emph{arXiv preprint arXiv:2307.03987}.

\bibitem[{Wang et~al.(2023{\natexlab{a}})Wang, Wang, Xu, Zhang, Gu, Jia, Yan, Zhang, and Sang}]{wang2023amber}
Junyang Wang, Yuhang Wang, Guohai Xu, Jing Zhang, Yukai Gu, Haitao Jia, Ming Yan, Ji~Zhang, and Jitao Sang. 2023{\natexlab{a}}.
\newblock Amber: An llm-free multi-dimensional benchmark for mllms hallucination evaluation.
\newblock \emph{arXiv preprint arXiv:2311.07397}.

\bibitem[{Wang et~al.(2023{\natexlab{b}})Wang, Zhao, Ouyang, Wang, and Shen}]{wang2023chatcad}
Sheng Wang, Zihao Zhao, Xi~Ouyang, Qian Wang, and Dinggang Shen. 2023{\natexlab{b}}.
\newblock Chatcad: Interactive computer-aided diagnosis on medical image using large language models.
\newblock \emph{arXiv preprint arXiv:2302.07257}.

\bibitem[{Wang et~al.(2024)Wang, Pan, Ding, and Biemann}]{wang2024mitigating}
Xintong Wang, Jingheng Pan, Liang Ding, and Chris Biemann. 2024.
\newblock Mitigating hallucinations in large vision-language models with instruction contrastive decoding.
\newblock \emph{arXiv preprint arXiv:2403.18715}.

\bibitem[{Woo et~al.(2024)Woo, Kim, Jang, Choi, and Kim}]{woo2024don}
Sangmin Woo, Donguk Kim, Jaehyuk Jang, Yubin Choi, and Changick Kim. 2024.
\newblock Don't miss the forest for the trees: Attentional vision calibration for large vision language models.
\newblock \emph{arXiv preprint arXiv:2405.17820}.

\bibitem[{Wu et~al.(2024)Wu, Liu, Wang, Zhang, Wu, Wang, and Tan}]{wu2024logical}
Junfei Wu, Qiang Liu, Ding Wang, Jinghao Zhang, Shu Wu, Liang Wang, and Tieniu Tan. 2024.
\newblock Logical closed loop: Uncovering object hallucinations in large vision-language models.
\newblock \emph{arXiv preprint arXiv:2402.11622}.

\bibitem[{Yin et~al.(2024)Yin, Fu, Zhao, Xu, Wang, Sui, Shen, Li, Sun, and Chen}]{yin2024woodpecker}
Shukang Yin, Chaoyou Fu, Sirui Zhao, Tong Xu, Hao Wang, Dianbo Sui, Yunhang Shen, Ke~Li, Xing Sun, and Enhong Chen. 2024.
\newblock Woodpecker: Hallucination correction for multimodal large language models.
\newblock \emph{Science China Information Sciences}, 67(12):220105.

\bibitem[{Yu et~al.(2024)Yu, Li, Wei, Pang, Ye, Qin, Tang, Tian, and Zhuang}]{yu2024hallucidoctor}
Qifan Yu, Juncheng Li, Longhui Wei, Liang Pang, Wentao Ye, Bosheng Qin, Siliang Tang, Qi~Tian, and Yueting Zhuang. 2024.
\newblock Hallucidoctor: Mitigating hallucinatory toxicity in visual instruction data.
\newblock In \emph{Proceedings of the IEEE/CVF Conference on Computer Vision and Pattern Recognition}, pages 12944--12953.

\bibitem[{Yuan et~al.(2024)Yuan, Cao, Jin, Chen, Zeng, Liu, and Zhao}]{yuan2024whispers}
Hongbang Yuan, Pengfei Cao, Zhuoran Jin, Yubo Chen, Daojian Zeng, Kang Liu, and Jun Zhao. 2024.
\newblock Whispers that shake foundations: Analyzing and mitigating false premise hallucinations in large language models.
\newblock \emph{arXiv preprint arXiv:2402.19103}.

\bibitem[{Zhang et~al.(2025)Zhang, Wan, Kan, Ma, Stepputtis, Ramanan, Salakhutdinov, Morency, Sycara, and Xie}]{zhang2025self}
Ce~Zhang, Zifu Wan, Zhehan Kan, Martin~Q Ma, Simon Stepputtis, Deva Ramanan, Russ Salakhutdinov, Louis-Philippe Morency, Katia Sycara, and Yaqi Xie. 2025.
\newblock Self-correcting decoding with generative feedback for mitigating hallucinations in large vision-language models.
\newblock \emph{arXiv preprint arXiv:2502.06130}.

\bibitem[{Zhou et~al.(2024)Zhou, Cui, Rafailov, Finn, and Yao}]{zhou2024aligning}
Yiyang Zhou, Chenhang Cui, Rafael Rafailov, Chelsea Finn, and Huaxiu Yao. 2024.
\newblock Aligning modalities in vision large language models via preference fine-tuning.
\newblock \emph{arXiv preprint arXiv:2402.11411}.

\bibitem[{Zhou et~al.(2023)Zhou, Cui, Yoon, Zhang, Deng, Finn, Bansal, and Yao}]{zhou2023analyzing}
Yiyang Zhou, Chenhang Cui, Jaehong Yoon, Linjun Zhang, Zhun Deng, Chelsea Finn, Mohit Bansal, and Huaxiu Yao. 2023.
\newblock Analyzing and mitigating object hallucination in large vision-language models.
\newblock \emph{arXiv preprint arXiv:2310.00754}.

\bibitem[{Zhu et~al.(2024)Zhu, Ji, Chen, Xu, Ye, and Liu}]{zhu2024ibd}
Lanyun Zhu, Deyi Ji, Tianrun Chen, Peng Xu, Jieping Ye, and Jun Liu. 2024.
\newblock Ibd: Alleviating hallucinations in large vision-language models via image-biased decoding.
\newblock \emph{arXiv preprint arXiv:2402.18476}.

\end{thebibliography}

\clearpage
\appendix

\section*{Appendix}

\section{More Implementation Details}
\subsection{Adaptive Plausibility Constraints}

Typically, current decoding methods employ adaptive plausibility constraints to calibrate the output distribution, truncating the generation of implausible outputs. The plausible vocabulary \( V_\text{head}(y_{<t}) \) is defined as:
\begin{equation}
  \begin{aligned}
    V_\text{head}(y_{<t}) = \{ y_t \in V : &p_\theta(y_t \mid v, x, y_{<t}) \geq \\
    &\beta \max_w p_\theta(w \mid v, x, y_{<t}) \},
  \end{aligned}
  \label{eq:head constrain}
\end{equation}
where \(V\) is the whole vocabulary of the LVLM, and \(\beta\) controls the strength of truncation. Any potential output \( y_t \) not residing within \( V_\text{head}(y_{<t}) \) is assigned a probability of zero:
\begin{equation}
    p_\theta(y_t \mid v, x, y_{<t}) = 0, \quad \text{if } y_t \notin V_\text{head}(y_{<t}).
\end{equation}
A larger \(\beta\) indicates more aggressive truncation, preserving only tokens with higher output probabilities. In our MoD framework, we set \(\beta = 0.5\) across all models and tasks.

\subsection{General Decoding Parameters}
To guarantee adequate generation diversity, we set \(temperature = 1\) and \(Top\text{-}p = 1\). Additionally, we configure \(max~new~tokens = 1024\), enabling the model to autonomously determine the appropriate stopping point for text generation, which fully reflects the model's capabilities.

\section{More Experimental Details}
\label{sec:exp details}

\subsection{Metrics}
\label{sec: metric details}
\noindent\textbf{Metrics on MME.}
In the image set \( \mathcal{I} \) of the MME benchmark, each image \( i \in \mathcal{I} \) is paired with a set of two similar questions designed to elicit ``yes'' and ``no'' responses, denoted as \( \{q_y^i, q_n^i\} \). To comprehensively evaluate the performance of models, two metrics are introduced: the question-level accuracy (Acc) and the image-level accuracy (Acc+). A correctly answered question contributes to Acc, while a correctly answered pair of questions for an image contributes to Acc+. Although Acc+ is a more rigorous metric, it provides a more comprehensive assessment of the model's ability to fully understand the images. The formulas for these metrics are defined as follows:
\begin{equation}
\begin{aligned}
    \text{Acc} = \Bigl\{&\sum_{i \in \mathcal{I}} \bigl[ \text{LVLM}(i, q_y^i) = \text{"Yes"} \bigr] \mathbin{+} \\ &\sum_{i \in \mathcal{I}} \bigl[ \text{LVLM}(i, q_n^i) = \text{"No"} \bigr]\Bigr\} \mathbin{/} \bigl(|\mathcal{I}| \times 2 \bigr),
\end{aligned}
\end{equation}
\begin{equation}
\begin{aligned}
    \text{Acc+} = \sum_{i \in \mathcal{I}} \bigl[ &\text{LVLM}(i, q_y^i) = \text{"Yes"} \mathbin{\land} \\ &\text{LVLM}(i, q_n^i) = \text{"No"} \bigr] \mathbin{/} |\mathcal{I}|,
\end{aligned}
\end{equation}
where \( \text{LVLM}(i, q_*^i) \) denotes the model's output when provided with image \( i \) and the corresponding question \( q_*^i \). The reported MME Score is computed as the sum of Acc and Acc+.

\noindent\textbf{Metrics on CHAIR.}
The CHAIR evaluation benchmark comprises two key metrics: \(\text{CHAIR}_i\) and \(\text{CHAIR}_s\), which measure hallucinations in image captions at the instance level and sentence level, respectively. Specifically, the instance-level metric \(\text{CHAIR}_i\) denotes the proportion of
hallucinated objects relative to all mentioned objects in the generated captions, and the sentence-level metric \(\text{CHAIR}_s\) reflects the proportion of generated
captions that contain hallucinations. In addition, we
assess the semantic completeness of generated captions with the metric Recall. Their formulas are defined as follows:
\begin{equation}
\text{CHAIR}_i = \frac{|\{ \text{hallucinated objects} \}|}{|\{ \text{all mentioned objects} \}|},
\label{eq:chair_i}
\end{equation}
\begin{equation}
\text{CHAIR}_s = \frac{|\{ \text{sentences with hallucinations} \}|}{|\{ \text{all sentences} \}|}.
\end{equation}
\begin{equation}
\text{Recall} = \frac{|\{ \text{accurately mentioned objects} \}|}{|\{ \text{ground-truth objects} \}|}.
\end{equation}

\noindent\textbf{Metrics on AMBER.}
For generative tasks in the AMBER benchmark, the official evaluation metrics include CHAIR (equivalent to $\text{CHAIR}_i$ in Eq.~\ref{eq:chair_i}, hence we use $\text{CHAIR}_i$ in the main text), Cover, Hal, and Cog. Let $R_{\text{obj}} = \{r_1, r_2, \dots, r_m\}$ denote the set of objects mentioned in the response generated by the LVLM, and $G_{\text{obj}} = \{g_1, g_2, \dots, g_n\}$ denote the set of ground truth objects annotated as present in the image. The specific calculation methods for each metric are as follows: \\
(1) \textbf{CHAIR} evaluates the frequency of hallucinated objects out of all mentioned objects in the model's response, and is defined as:
\begin{equation}
\text{CHAIR} = 1 - \frac{|R_{\text{obj}} \cap G_{\text{obj}}|}{|R_{\text{obj}}|}.
\end{equation}
(2) \textbf{Cover} assesses the degree to which ground-truth objects are accurately captured within the model's generated response. Specifically, it quantifies the proportion of ground-truth objects that are faithfully represented in the output. The formula for Cover is defined as follows: 
\begin{equation}
\text{Cover} = \frac{|R_{\text{obj}} \cap G_{\text{obj}}|}{|G_{\text{obj}}|}.
\end{equation}
(3) \textbf{Hal} quantifies the proportion of hallucinated responses relative to the total responses generated by the model, computed as follows:
\begin{equation}
\text{Hal} = 
\begin{cases} 
1, & \text{if } \text{CHAIR} \neq 0 \\
0, & \text{otherwise}
\end{cases}.
\end{equation}
(4) \textbf{Cog} assesses the similarity between the hallucinations generated by LVLMs and those commonly observed in human responses. To this end, the authors annotated a set of objects $H_{\text{obj}} = \{h_1, h_2, \dots, h_p\}$ that humans are prone to hallucinate. Cog is calculated as:
\begin{equation}
\text{Cog} = \frac{|R_{\text{obj}} \cap H_{\text{obj}}|}{|R_{\text{obj}}|}.
\end{equation}
The final evaluation scores are obtained by computing the average values of these four metrics across all queries in the AMBER benchmark. 

In our experiments, we employ the following versions of the libraries and models: NLTK version 3.9.1, SpaCy version 2.3.9, and the large English language model en\_core\_web\_lg, version 2.3.0.

\subsection{Licensing}
\textbf{Benchmarks license.}
POPE~\cite{li2023evaluating} is licensed under the MIT License. MME~\cite{fu2023mme} is restricted for academic research purposes only. CHAIR~\cite{rohrbach2018object}  is distributed under the BSD 2-Clause License. AMBER~\cite{wang2023amber} is released under the CC0 license.

\noindent\textbf{LVLMs license.}
LLaVA-1.5~\cite{liu2024llava} and LLaVA-NEXT~\cite{liu2024llavanext} are licensed under the Apache-2.0 License. Qwen-VL~\cite{bai2023qwenvl} permits academic use.

\noindent\textbf{Baselines license.}
VCD~\cite{leng2024mitigating} is licensed under the Apache-2.0 License. M3ID~\cite{favero2024multi} is available for academic use. AvisC~\cite{woo2024don} is licensed under MIT License.

\subsection{Results}

\begin{table*}
    \centering
    \resizebox{\linewidth}{!}{
    \begin{tabular}{l l l @{\hspace{25pt}} c c c c @{\hspace{25pt}} c c c c @{\hspace{25pt}} c c c c}
    \toprule
        \multirow{2}{*}{\textbf{Dataset}} & \multirow{2}{*}{\textbf{Setting}} & \multirow{2}{*}{\textbf{Method}} & \multicolumn{4}{c@{\hspace{25pt}}}{\textbf{LLaVA-v1.5}} & \multicolumn{4}{c@{\hspace{25pt}}}{\textbf{Qwen-VL}} & \multicolumn{4}{c@{\hspace{15pt}}}{\textbf{LLaVA-NEXT}} \\
        ~ & ~ & ~ & Acc & Pre & Rec & F1 & Acc & Pre & Rec & F1 & Acc & Pre & Rec & F1 \\
        \midrule
        \multirow{15}{*}{MS-COCO} & \multirow{5}{*}{random} & sampling & 83.8 & 82.4 & 86.1 & 84.2 & 84.9 & 96.0 & 72.9 & 82.9 & 84.4 & 94.7 & 72.8 & 82.3 \\
        ~ & ~ & VCD & 85.0 & 82.7 & 86.1 & 84.2 & 85.5 & 96.0 & 71.1 & 83.6 & 86.0 & 96.5 & 74.8 & 84.3 \\
        ~ & ~ & M3ID & 86.1 & 83.2 & 86.8 & 85.0 & 85.3 & 95.1 & 74.2 & 83.4 & 85.5 & 96.3 & 73.9 & 83.6 \\
        ~ & ~ & AvisC & 82.3 & 78.4 & 89.3 & 83.5 & 82.9 & 96.2 & 68.5 & 80.0 & 85.2 & 98.6 & 71.3 & 82.8 \\
        ~ & ~ & MoD (Ours) & \textbf{89.2} & 90.0 & 88.2 & \textbf{89.1} & \textbf{86.0} & 97.0 & 74.3 & \textbf{84.1} & \textbf{86.6} & 97.7 & 74.9 & \textbf{84.8} \\
        \cmidrule{2-15}
        ~ & \multirow{5}{*}{popular} & sampling & 82.0 & 79.7 & 85.9 & 82.6 & 84.0 & 94.7 & 72.1 & 81.9 & 83.2 & 90.9 & 73.8 & 81.5 \\
        ~ & ~ & VCD & 82.1 & 78.5 & 88.3 & 83.2 & 84.9 & 94.5 & 74.9 & 83.6 & 84.5 & 92.9 & 74.8 & 82.9 \\
        ~ & ~ & M3ID & 82.8 & 80.1 & 88.5 & 84.1 & 84.2 & 94.1 & 73.8 & 82.7 & 84.2 & 93.1 & 73.9 & 82.4 \\
        ~ & ~ & AvisC & 78.2 & 72.7 & 90.3 & 80.5 & 82.8 & 95.5 & 68.9 & 80.1 & 83.9 & 94.7 & 71.7 & 81.6 \\
        ~ & ~ & MoD (Ours) & \textbf{85.7} & 84.1 & 88.1 & \textbf{86.1} & \textbf{85.6} & 96.3 & 74.0 & \textbf{83.7} & \textbf{85.5} & 95.1 & 74.9 & \textbf{83.8} \\
        \cmidrule{2-15}
        ~ & \multirow{5}{*}{adversarial} & sampling & 75.8 & 71.3 & 86.3 & 78.1 & 82.1 & 90.0 & 72.3 & 80.2 & 79.5 & 84.1 & 72.9 & 78.1 \\
        ~ & ~ & VCD & 76.3 & 71.5 & 87.3 & 78.7 & 84.0 & 90.6 & 74.9 & 82.0 & 80.9 & 85.2 & 74.8 & 79.7 \\
        ~ & ~ & M3ID & 77.1 & 71.8 & 87.6 & 78.9 & 83.2 & 90.4 & 73.1 & 80.8 & 80.6 & 85.4 & 73.9 & 79.2 \\
        ~ & ~ & AvisC & 74.2 & 68.4 & 89.9 & 77.7 & 81.2 & 91.9 & 68.5 & 78.5 & 81.8 & 91.3 & 71.5 & 80.2 \\
        ~ & ~ & MoD (Ours) & \textbf{79.7} & 75.4 & 88.2 & \textbf{81.3} & \textbf{84.0} & 92.4 & 74.2 & \textbf{82.3} & \textbf{82.4} & 88.3 & 74.8 & \textbf{81.0} \\
        \midrule
        \multirow{15}{*}{A-OKVQA} & \multirow{5}{*}{random} & sampling & 81.8 & 76.4 & 92.1 & 83.5 & 86.8 & 93.2 & 79.5 & 85.8 & 83.8 & 87.2 & 79.2 & 83.0 \\
        ~ & ~ & VCD & 81.2 & 75.2 & 93.0 & 83.2 & 87.4 & 92.9 & 81.1 & 86.6 & 84.8 & 89.2 & 79.3 & 83.9 \\
        ~ & ~ & M3ID & 82.9 & 76.8 & 94.1 & 84.6 & 87.1 & 92.4 & 80.3 & 85.9 & 85.3 & 90.7 & 78.7 & 84.3 \\
        ~ & ~ & AvisC & 79.1 & 71.9 & 95.5 & 82.1 & 84.7 & 93.0 & 74.9 & 83.0 & 85.8 & 93.4 & 76.9 & 84.4 \\
        ~ & ~ & MoD (Ours) & \textbf{86.5} & 81.3 & 94.7 & \textbf{87.5} & \textbf{87.8} & 94.4 & 80.4 & \textbf{86.9} & \textbf{86.4} & 91.0 & 80.7 & \textbf{85.6} \\
        \cmidrule{2-15}
        ~ & \multirow{5}{*}{popular} & sampling & 75.3 & 69.1 & 91.5 & 78.7 & 85.6 & 90.6 & 79.5 & 84.7 & 81.4 & 83.4 & 78.3 & 80.8 \\
        ~ & ~ & VCD & 74.7 & 68.2 & 92.5 & 78.5 & 86.3 & 89.5 & 81.2 & 85.1 & 81.5 & 82.6 & 79.9 & 81.2 \\
        ~ & ~ & M3ID & 75.8 & 69.8 & 92.1 & 79.4 & 85.9 & 90.7 & 79.2 & 84.6 & 82.2 & 84.7 & 78.7 & 81.6 \\
        ~ & ~ & AvisC & 71.8 & 64.7 & 95.6 & 77.2 & 83.9 & 90.9 & 75.5 & 82.4 & 83.9 & 90.1 & 76.5 & 82.8 \\
        ~ & ~ & MoD (Ours) & \textbf{79.5} & 72.6 & 94.5 & \textbf{82.2} & \textbf{86.5} & 91.6 & 80.3 & \textbf{85.6} & \textbf{84.2} & 86.7 & 80.7 & \textbf{83.6} \\
        \cmidrule{2-15}
        ~ & \multirow{5}{*}{adversarial} & sampling & 67.4 & 61.8 & 91.2 & 73.7 & 80.4 & 80.1 & 80.9 & 80.5 & 73.2 & 71.0 & 78.4 & 74.5 \\
        ~ & ~ & VCD & 68.1 & 61.9 & 93.8 & 74.6 & 80.7 & 80.1 & 81.6 & 80.8 & 74.7 & 72.2 & 80.3 & 76.0 \\
        ~ & ~ & M3ID & 68.3 & 62.1 & 93.4 & 74.6 & 80.5 & 80.7 & 80.2 & 80.4 & 74.6 & 72.5 & 79.3 & 75.7 \\
        ~ & ~ & AvisC & 64.4 & 58.8 & 96.1 & 73.0 & 78.1 & 80.1 & 74.7 & 77.3 & 75.6 & 76.5 & 76.7 & 76.6 \\
        ~ & ~ & MoD (Ours) & \textbf{69.1} & 62.6 & 94.7 & \textbf{75.4} & \textbf{81.0} & 81.2 & 80.6 & \textbf{80.9} & \textbf{75.7} & 73.3 & 80.9 & \textbf{76.9} \\
        \midrule
        \multirow{15}{*}{GQA} & \multirow{5}{*}{random} & sampling & 81.6 & 75.6 & 93.2 & 83.5 & 81.3 & 88.8 & 71.5 & 79.2 & 83.1 & 85.8 & 79.4 & 82.5 \\
        ~ & ~ & VCD & 82.2 & 76.0 & 94.1 & 84.1 & 82.0 & 87.6 & 74.5 & 80.5 & 84.0 & 87.0 & 80.1 & 83.4\\
        ~ & ~ & M3ID & 83.3 & 76.8 & 94.0 & 84.5 & 82.4 & 88.1 & 72.8 & 79.7 & 84.3 & 88.2 & 79.3 & 83.5 \\
        ~ & ~ & AvisC & 79.0 & 71.4 & 96.7 & 82.2 & 80.5 & 89.9 & 68.6 & 77.8 & 84.9 & 92.7 & 75.9 & 83.4 \\
        ~ & ~ & MoD (Ours) & \textbf{86.2} & 80.8 & 95.1 & \textbf{87.4} & \textbf{83.8} & 90.3 & 75.7 & \textbf{82.3} & \textbf{86.2} & 90.3 & 81.2 & \textbf{85.5} \\
        \cmidrule{2-15}
        ~ & \multirow{5}{*}{popular} & sampling & 73.1 & 66.7 & 92.5 & 77.5 & 75.9 & 78.1 & 72.0 & 74.9 & 78.5 & 78.7 & 78.2 & 78.5 \\
        ~ & ~ & VCD & 71.5 & 64.7 & 94.5 & 76.8 & 75.9 & 76.6 & 74.7 & 75.6 & 78.2 & 77.2 & 80.1 & 78.6 \\
        ~ & ~ & M3ID & 72.3 & 64.9 & 94.8 & 77.1 & 76.8 & 78.9 & 75.2 & 77.0 & 78.8 & 78.5 & 79.3 & 78.9 \\
        ~ & ~ & AvisC & 67.4 & 60.9 & 97.1 & 74.8 & 74.2 & 77.9 & 67.5 & 72.3 & 80.5 & 83.8 & 75.6 & 79.5 \\
        ~ & ~ & MoD (Ours) & \textbf{74.0} & 66.8 & 95.3 & \textbf{78.6} & \textbf{79.8} & 82.4 & 75.7 & \textbf{78.9} & \textbf{81.1} & 81.0 & 81.2 & \textbf{81.1} \\
        \cmidrule{2-15}
        ~ & \multirow{5}{*}{adversarial} & sampling & 68.0 & 62.0 & 93.4 & 74.5 & 75.5 & 77.8 & 71.2 & 74.4 & 73.3 & 71.3 & 78.0 & 74.5 \\
        ~ & ~ & VCD & 67.6 & 61.5 & 94.4 & 74.5 & 76.7 & 77.8 & 74.7 & 76.2 & 74.2 & 71.8 & 79.8 & 75.6 \\
        ~ & ~ & M3ID & 67.2 & 61.0 & 93.9 & 74.0 & 77.1 & 78.6 & 74.9 & 76.7 & 74.2 & 72.1 & 79.1 & 75.4 \\
        ~ & ~ & AvisC & 64.1 & 58.5 & 96.7 & 72.9 & 75.5 & 80.5 & 67.4 & 73.4 & 75.8 & 76.2 & 75.6 & 75.9 \\
        ~ & ~ & MoD (Ours) & \textbf{68.7} & 62.2 & 95.3 & \textbf{75.3} & \textbf{78.9} & 81.0 & 75.4 & \textbf{78.1} & \textbf{76.0} & 73.6 & 81.2 & \textbf{77.2} \\
    \bottomrule
    \end{tabular}
    }
    \caption{\textbf{Detailed evaluation results on POPE benchmark.} ``Acc'', ``Pre'', ``Rec'', and ``F1'' stand for Accuracy, Precision, Recall, and F1 score, respectively.}
    \label{tab:POPE detailed result}
\end{table*}

\label{sec:detailed pope}
\noindent\textbf{Detailed results on POPE.}
We conduct a comprehensive evaluation of MoD on the POPE benchmark across three datasets: MS-COCO, A-OKVQA, and GQA. POPE consists of 500 images from each dataset, with each image associated with six probing questions, resulting in a total of 27,000 query-answer pairs. As detailed in Tab.~\ref{tab:POPE detailed result}, MoD demonstrates consistent and superior performance compared to other decoding methods across all 27 experimental settings. Notably, MoD achieves substantial improvements in both Accuracy and F1 score compared to the second-best methods, with maximum performance gains of 3.7 and 4.1 points for LLaVA-1.5, 3.0 and 1.9 points for Qwen-VL, and 1.3 and 2.0 points for LLaVA-NEXT, highlighting its effectiveness in reducing hallucinations and generating reliable outputs.

\subsection{Ablation Studies}
\label{sec: other ablas}
As previously discussed, MoD distinguishes itself from other decoding strategies by employing a shared set of hyperparameters across diverse tasks and model architectures. Furthermore, MoD exhibits low sensitivity to variations in hyperparameters, as detailed in the following analysis.

\begin{figure}
  \includegraphics[width=\columnwidth]{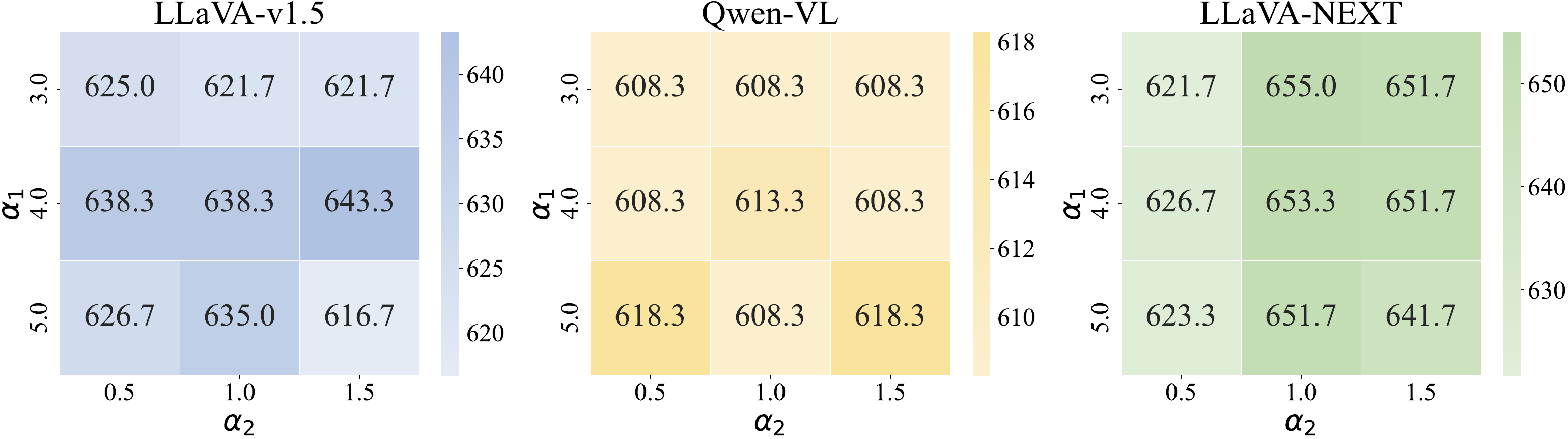}
  \caption{\textbf{Ablation study of the decoding coefficient \(\alpha_1\) and \(\alpha_2\)} on MME benchmark. We set \(\alpha_1 = 4.0\) and \(\alpha_2 = 1.0\) for all three evaluated LVLMs. For reference, the optimal performances achieved by the baseline methods in our experiments were 596.7, 593.3, and 613.3 for each respective model.}
  \label{fig:alpha ablation}
\end{figure}

\noindent\textbf{Analysis of the decoding coefficient \(\alpha_1\) and \(\alpha_2\).}
In Fig.~\ref{fig:alpha ablation}, we present the results of the ablation study on the impact of the complementary decoding coefficient \(\alpha_1\) and the contrastive decoding coefficient \(\alpha_2\) in Eq.~\ref{eq:MoD} on MME benchmark. The total MME Score is evaluated by simultaneously varying \(\alpha_1\) within the range of 3.0 to 5.0 and \(\alpha_2\) within the range of 0.5 to 1.5, and we observe that MoD exhibits highly stable performance across the tested parameter configurations. Specifically, the maximum deviations are 3.4\% for LLaVA-1.5, 0.8\% for Qwen-VL, and 4.8\% for LLaVA-NEXT, compared to the results obtained using our chosen configuration, indicating the robustness of MoD with respect to variations in \(\alpha_1\) and \(\alpha_2\).

\noindent\textbf{Analysis of the adaptive plausible constraints coefficient \(\beta\).}
In Tab.~\ref{tab:beta ablation}, we conduct an ablation study to investigate the impact of the adaptive plausible constraints coefficient \(\beta\) in Eq.~\ref{eq:head constrain} on the MME benchmark. By varying \(\beta\) within the range of 0.3 to 0.7, we observe that the total MME Score exhibits remarkable stability across different models. Specifically, the maximum deviations are only 1.8\% for LLaVA-1.5, 0.8\% for Qwen-VL, and 3.1\% for LLaVA-NEXT when compared to the results obtained using our selected \(\beta\) value. Notably, even the least favorable results still surpass the best baseline methods, underscoring the robustness of our approach concerning the choice of \(\beta\).

\noindent\textbf{Analysis of the proportion \(\lambda\) for selecting attended image tokens.}
In Tab.~\ref{tab:lambda ablation}, we conduct an ablation study to evaluate the influence of the proportion \(\lambda\) for selecting attended image tokens defined in Eq.~\ref{eq:high attn select}. Experimental results demonstrate that MoD achieves remarkable stability across a wide range of \(\lambda\) values, from 0.10 to 0.30. Notably, we observe a more pronounced degradation in performance when \(\lambda\) is decreased compared to when it is increased from 0.20. This asymmetry can be attributed to the fact that decreasing \(\lambda\) leads to a diminished amount of visual information for both complementary and contrastive strategies, thereby affecting the overall performance.

\begin{table}
    \centering
    \resizebox{\columnwidth}{!}{
    \begin{tabular}{l c c c c c c}
    \toprule
        \multirow{2}{*}{\textbf{Model}} & \multirow{2}{*}{\textbf{\phantom{*}$\beta$\phantom{*}}} & \multicolumn{2}{c}{\textbf{Object-level}} & \multicolumn{2}{c}{\textbf{Attribute-level}} & \multirow{2}{*}{\textbf{Total}} \\ 
        ~ & ~ & Existence & Count & Position & Color & ~ \\
        \midrule
        \multirow{5}{*}{LLaVA-v1.5} & \phantom{*}0.3\phantom{*} & 195.0 & 146.7 & 131.7 & 160.0 & 633.3 \\
        ~ & \phantom{*}0.4\phantom{*} & 195.0 & 151.7 & 131.7 & 155.0 & 633.3 \\
        ~ & \phantom{*}0.5* & 195.0 & 141.7 & 126.7 & 175.0 & \textbf{638.3} \\
        ~ & \phantom{*}0.6\phantom{*} & 195.0 & 145.0 & 138.3 & 155.0 & 633.3 \\
        ~ & \phantom{*}0.7\phantom{*} & 195.0 & 143.3 & 133.3 & 155.0 & 626.7 \\
        \midrule
        \multirow{5}{*}{Qwen-VL} & \phantom{*}0.3\phantom{*} & 160.0 & 160.0 & 108.3 & 180.0 & 608.3 \\
        ~ & \phantom{*}0.4\phantom{*} & 165.0 & 160.0 & 108.3 & 185.0 & \textbf{618.3} \\
        ~ & \phantom{*}0.5* & 170.0 & 160.0 & 103.3 & 180.0 & 613.3 \\
        ~ & \phantom{*}0.6\phantom{*} & 170.0 & 155.0 & 103.3 & 180.0 & 608.3 \\
        ~ & \phantom{*}0.7\phantom{*} & 165.0 & 155.0 & 103.3 & 185.0 & 608.3 \\
        \midrule
        \multirow{5}{*}{LLaVA-NEXT} & \phantom{*}0.3\phantom{*} & 195.0 & 146.7 & 126.7 & 165.0 & 633.3 \\
        ~ & \phantom{*}0.4\phantom{*} & 195.0 & 155.0 & 121.7 & 170.0 & 641.7 \\
        ~ & \phantom{*}0.5* & 195.0 & 160.0 & 133.3 & 165.0 & \textbf{653.3} \\
        ~ & \phantom{*}0.6\phantom{*} & 195.0 & 146.7 & 143.3 & 160.0 & 645.0 \\
        ~ & \phantom{*}0.7\phantom{*} & 195.0 & 160.0 & 138.3 & 155.0 & 648.3 \\
    \bottomrule
    \end{tabular}
    }
    \caption{\textbf{Ablation study of the adaptive plausible constraints coefficient $\beta$} on MME benchmark. For reference, the best total MME Score achieved by the baseline methods in our experiments were 596.7, 593.3, and 613.3 for each respective model. *We choose \(\beta = 0.5\) for all three evaluated LVLMs.}
    \label{tab:beta ablation}
\end{table}

\begin{table}
    \centering
    \resizebox{\columnwidth}{!}{
    \begin{tabular}{l c c c c c c}
    \toprule
        \multirow{2}{*}{\textbf{Model}} & \multirow{2}{*}{\textbf{\phantom{*}$\lambda$\phantom{*}}} & \multicolumn{2}{c}{\textbf{Object-level}} & \multicolumn{2}{c}{\textbf{Attribute-level}} & \multirow{2}{*}{\textbf{Total}} \\ 
        ~ & ~ & Existence & Count & Position & Color & ~ \\
        \midrule
        \multirow{5}{*}{LLaVA-v1.5} & \phantom{*}0.10\phantom{*} & 195.0 & 138.3 & 131.7 & 150.0 & 615.0 \\
        ~ & \phantom{*}0.15\phantom{*} & 195.0 & 140.0 & 126.7 & 155.0 & 616.7 \\
        ~ & \phantom{*}0.20* & 195.0 & 141.7 & 126.7 & 175.0 & 638.3 \\
        ~ & \phantom{*}0.25\phantom{*} & 195.0 & 160.0 & 118.3 & 160.0 & \textbf{633.3} \\
        ~ & \phantom{*}0.30\phantom{*} & 190.0 & 163.3 & 128.3 & 150.0 & 631.7 \\
        \midrule
        \multirow{5}{*}{Qwen-VL} & \phantom{*}0.10\phantom{*} & 170.0 & 155.0 & 103.3 & 175.0 & 603.3 \\
        ~ & \phantom{*}0.15\phantom{*} & 165.0 & 160.0 & 103.3 & 175.0 & 603.3 \\
        ~ & \phantom{*}0.20* & 170.0 & 160.0 & 103.3 & 180.0 & 613.3 \\
        ~ & \phantom{*}0.25\phantom{*} & 170.0 & 165.0 & 103.3 & 180.0 & \textbf{618.3} \\
        ~ & \phantom{*}0.30\phantom{*} & 165.0 & 150.0 & 108.3 & 180.0 & 603.3 \\
        \midrule
        \multirow{5}{*}{LLaVA-NEXT} & \phantom{*}0.10\phantom{*} & 190.0 & 155.0 & 93.3 & 170.0 & 608.3 \\
        ~ & \phantom{*}0.15\phantom{*} & 195.0 & 153.3 & 120.0 & 170.0 & 638.3 \\
        ~ & \phantom{*}0.20* & 195.0 & 160.0 & 133.3 & 165.0 & \textbf{653.3} \\
        ~ & \phantom{*}0.25\phantom{*} & 195.0 & 138.3 & 116.7 & 180.0 & 630.0 \\
        ~ & \phantom{*}0.30\phantom{*} & 195.0 & 146.7 & 128.3 & 175.0 & 645.0 \\
    \bottomrule
    \end{tabular}
    }
    \caption{\textbf{Ablation study of the proportion \(\lambda\) for selecting attended image tokens} on MME benchmark. For reference, the best total MME Score achieved by the baseline methods in our experiments were 596.7, 593.3, and 613.3 for each respective model. *We choose \(\lambda = 0.20\) for all three evaluated LVLMs.}
    \label{tab:lambda ablation}
\end{table}

\subsection{Comparison with DeGF}
As an adaptive decoding method, we compared MoD with DeGF~\cite{zhang2025self} on MME. DeGF first generates an image based on the original output, then adaptively selects a contrastive decoding strategy by evaluating the consistency between the generated image's output and the original output. In contrast, MoD adaptively chooses its contrastive decoding strategy based on the consistency between the original output and the output derived from only partial image tokens. Experimental results show that DeGF slightly outperforms MoD overall, with each method having its own advantages in different subcategories. However, DeGF requires an additional inference step and an image generation process compared to MoD. Therefore, in practice, the choice of decoding strategy should balance computational cost and performance based on specific needs.
\begin{table}[ht]
    \centering
    \resizebox{\columnwidth}{!}{
    \begin{tabular}{l c c c c c}
    \toprule
        \multirow{2}{*}{\textbf{Method}} & \multicolumn{2}{c}{\textbf{Object-level}} & \multicolumn{2}{c}{\textbf{Attribute-level}} & \multirow{2}{*}{\textbf{Total}} \\ 
        ~ & Existence & Count & Position & Color & ~ \\
        \midrule
        sampling & 170.0 & 103.3 & 108.3 & 128.3 & 510.0 \\
        DeGF & 190.0 & 150.0 & 133.3 & 170.0 & 643.3 \\
        MoD & 195.0 & 141.7 & 126.7 & 175.0 & 638.3 \\
    \bottomrule
    \end{tabular}
    }
    \caption{\textbf{Comparison with DeGF using LLaVA-v1.5.}}
    \label{tab:degf}
\end{table}

\section{More Case Studies}
\label{sec: more cases}
\begin{figure}
  \includegraphics[width=\linewidth]{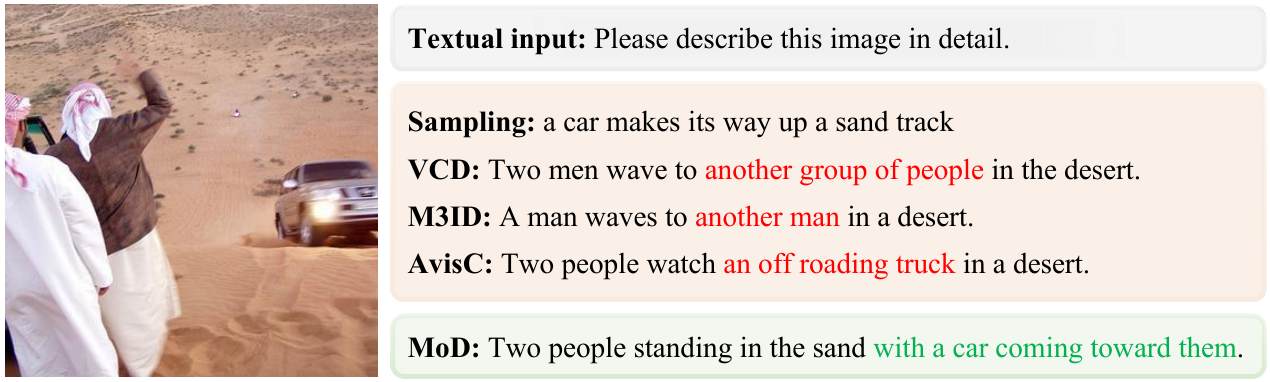}
  \caption {\textbf{Case study of generative tasks using Qwen-VL.} We compare responses generated by sampling, VCD, M3ID, AvisC, and our proposed MoD. Hallucinated content is highlighted in 
\textcolor[HTML]{ff0000}{red}, while more detailed and accurate content is marked in 
\textcolor[HTML]{06b050}{green}.}
  \label{fig:casestudy3}
\end{figure}

\begin{figure*}
  \includegraphics[width=\linewidth]{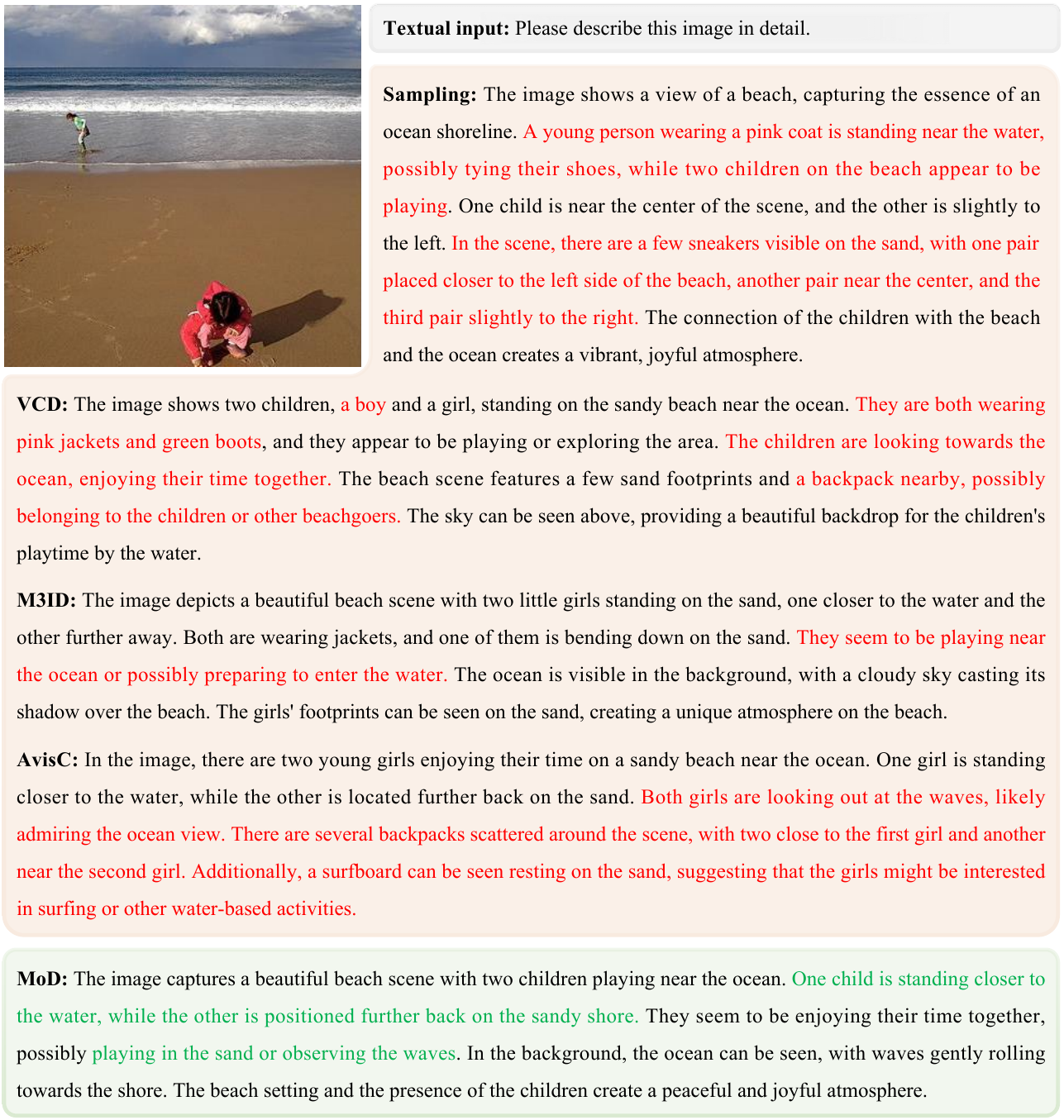}
  \caption {\textbf{Case study of generative tasks using LLaVA-1.5.} We compare responses generated by sampling, VCD, M3ID, AvisC, and our proposed MoD. Hallucinated content is highlighted in 
\textcolor[HTML]{ff0000}{red}, while more detailed and accurate content is marked in 
\textcolor[HTML]{06b050}{green}.}
  \label{fig:casestudy4}
\end{figure*}

\begin{figure*}
  \includegraphics[width=\linewidth]{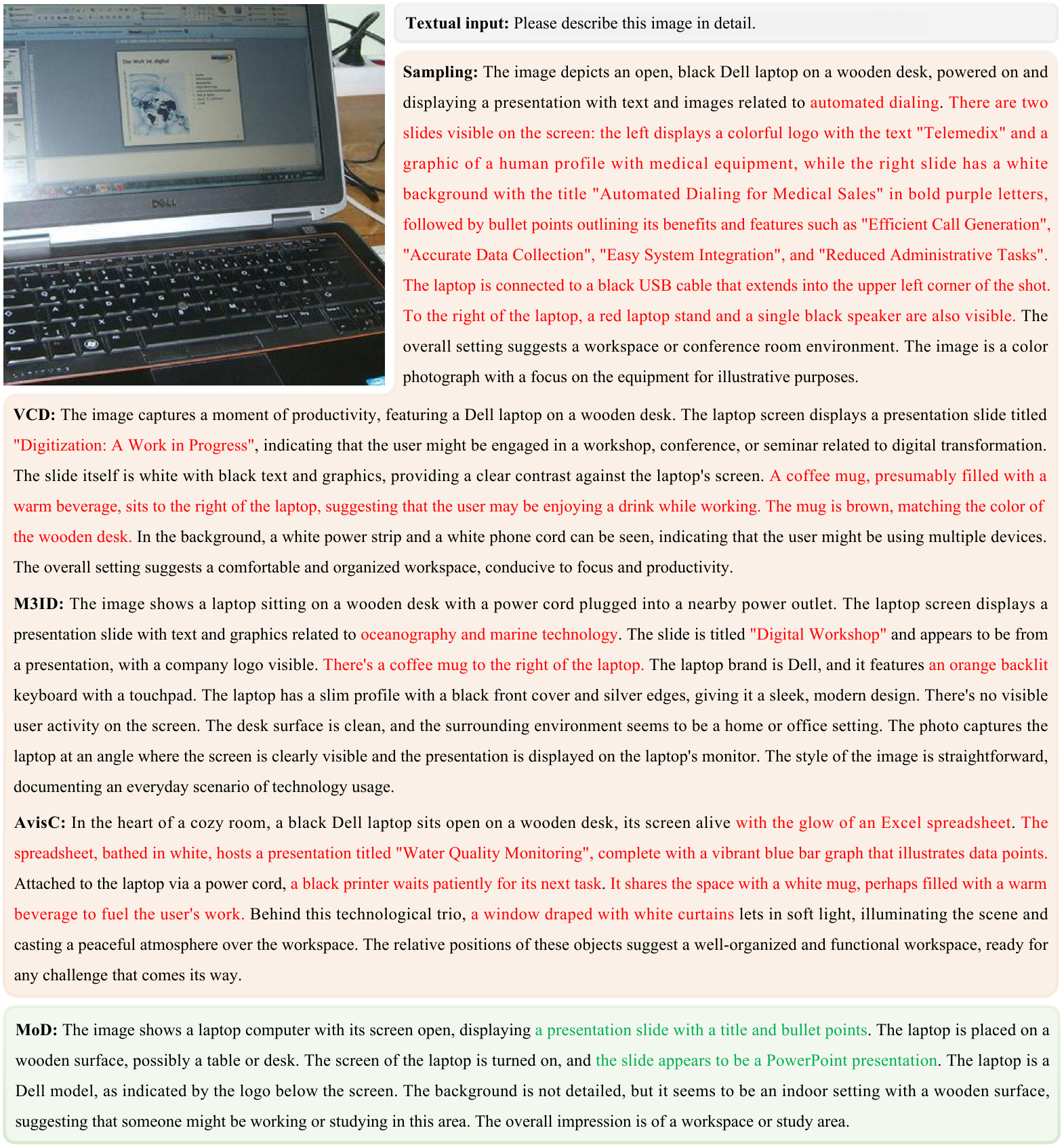}
  \caption {\textbf{Case study of generative tasks using LLaVA-NEXT.} We compare responses generated by sampling, VCD, M3ID, AvisC, and our proposed MoD. Hallucinated content is highlighted in 
\textcolor[HTML]{ff0000}{red}, while more detailed and accurate content is marked in 
\textcolor[HTML]{06b050}{green}.}
  \label{fig:casestudy5}
\end{figure*}

We provide additional case studies of generative tasks involving all three evaluated LVLMs, which generate responses of varying lengths. As shown in Fig.~\ref{fig:casestudy3}, MoD accurately identifies that the car is coming toward the two people. In Fig.~\ref{fig:casestudy4}, other decoding methods exhibit severe hallucinations, such as referring to ``a few sneakers'' or ``a backpack nearby''. Conversely, MoD effectively mitigates these hallucinations and provides an exact description of the girls’ location and orientation. In Fig.\ref{fig:casestudy5}, baseline methods tend to generate speculative content regarding the computer screen display, despite inadequate image resolution. These methods are also prone to hallucinate objects that commonly co-occur with computers, like ``coffee mug'', ``printer'' and so on. In contrast, MoD accurately generates an overview of the screen instead, without introducing any hallucinations, demonstrating its superior ability to generate accurate and detailed captions.

\end{document}